\documentclass[a4paper, 11pt, oneside]{Thesis}  
\usepackage{graphicx}
\graphicspath{{Figures/}}
\DeclareGraphicsExtensions{.pdf,.png,.jpg}

\usepackage[square, numbers, comma, sort&compress]{natbib}  
\usepackage{verbatim}  
\usepackage{vector}  
\hypersetup{urlcolor=black, 
            colorlinks=true} 

\usepackage{amsmath} 
\usepackage{mathrsfs}
\usepackage{algorithm}
\usepackage{multirow}
\usepackage{tablefootnote}
\usepackage{algpseudocode}
\usepackage{graphicx}
\usepackage{caption}

\begin{document}
\frontmatter      

\UNIVERSITY{{THE UNIVERSITY OF MELBOURNE }}    
%
\department{{Faculty of Science}}
\school{{School of Mathematics and Statistics}}


%
\title  {Synthesizing Tabular Data Using Selectivity Enhanced Generative Adversarial Networks}
\authors  {\texorpdfstring
            {\href{youran@student.unimelb.edu.au}{Youran Zhou}}
            {Author Name}
            }

\supervisor {Dr. Jianzhong Qi}

\addresses  {\groupname\\\deptname\\\univname}  
\date       {\today}
\subject    {}

\maketitle

\setstretch{1.3}  

\fancyhead{}  
\rhead{\thepage}  
\lhead{}  

\pagestyle{fancy}  


\addtotoc{Abstract}  
\abstract{
\addtocontents{toc}{\vspace{1em}}  

While the fast pace of economic development, E-commerce platforms face significant challenges in handling excessive customer transactions during major online shopping events like Black Friday. To be prepared for large volumes of transactions, those platforms need to utilize synthesized data to run stress tests and derive the computational resources needed to cope with such transactions.
The synthesized data for such patterns are usually in the form of tables. 

Generating Adversarial Networks (GAN) are used in most recent tabular data synthesizing studies and have shown impressive performance in generating tabular data while fulfilling privacy constraints and downstream machine learning model training needs. However, existing studies do not apply to the 
E-commerce stress testing scenarios directly because the computational resources required to process the data generated by GAN have not been considered. 
A core concept in computational resource estimation for database transaction processing is query selectivity. To the best of our knowledge, no study has been conducted on supporting selectivity constraints in the tabular data synthesizing field.
 
This thesis considers query selectivity constraints in tabular data generation and offers solutions by designing a novel method for tabular generation GAN models. We add a pre-trained deep neural network component for an additional supervision signal to model the query selectivity constraint that maintains the selectivity consistency between ground truth data and synthetic data. We implement our method on top of two GAN models 
and evaluate them with extensive experiments against the three state-of-the-art GAN models and a VAE model on five real-world datasets. The results show that the synthetic data generated by our model resembles the real data, increasing the selectivity estimation accuracy by up to 20\% and machine learning utilities by up to 6\%.

}

\clearpage  

%
%
\Declaration{

\addtocontents{toc}{\vspace{1em}}  


 


 I certify that this report does not incorporate without acknowledgement any material previously submitted for a degree or diploma in any university; and that to the best of my knowledge and belief it does not contain any material previously published or written by another person where due reference is not made in the text. The report is 11460 words in length (excluding text in images, tables, bibliographies and appendices).
 \\\\\\\\\\\
Signed: Youran Zhou\\
\rule[1em]{25em}{0.5pt}  
 
Date: 05/06/2022\\
\rule[1em]{25em}{0.5pt}  
}
\clearpage  


\setstretch{1.3}  
\acknowledgements{
\addtocontents{toc}{\vspace{1em}}  

I would like to thank our supervisors, Dr Jianzhong Qi as well as Dr Wei Wang from the Hong Kong University of Science and Technology, for your guidance and constant patience and encouragement throughout the year. As I look back, this precious experience with you has been the highlight of my Master's program. When I struggled, you always pointed me in the right direction and supported me during our weekly meetings. 

Thank you to Dr Qi for leading me to this research topic and showing me a different world I've never seen before. Preparing papers for me, scheduling our meetings, and helping me understand our project. Thank you for guiding my codes and experiments. Thank you for bearing my writing skills, providing detailed feedback and helping me with my thesis. It is my pleasure to have a great supervisor like you for my research project. I did grow and learned a lot from the past year. I am truly grateful for your kind words and encouragement.

Thank you to all kind staff from Spartan and IT support from the University of Melbourne for fixing my slurms and teaching me how to use the Spartan properly. Without you, I could not finish all of my experiments.

Lastly, I would like to thank my parents, my twin sister and our family 
mascot Jinzhi for sending me videos and voice calls to cheer me up, supporting and believing in me unconditionally. 
}

\clearpage  

\pagestyle{fancy}  

\lhead{\emph{Contents}}  
\tableofcontents  


\lhead{\emph{List of Figures}}  
\listoffigures  

\lhead{\emph{List of Tables}}  
\listoftables  






\mainmatter	  
\pagestyle{fancy}  

\lhead{\emph{Chapter 1 Introduction}}

\chapter{Introduction}
\label{chap: Introduction}

Back in 2017, The Economist published a story titled,  `The world's most valuable resource is no longer oil, but data.' Companies from a variety of industry fields gain valuable insights from using internal and external data sources. The desire of data raises the issue of data shortage. For instance, in the medical field, new technologies are utilizing patient health histories to create predictive models that can be used to improve diagnosis and understanding of illness. Rare diseases are challenging to study since we can only find a limited number of real-life datasets. In the E-commerce field, the online shopping platforms face the challenge from gigantic amount of transaction data during the major shopping events such as Black Friday, where a lack of computation resources may result in a blockage of user transactions, thus negatively impacting the revenue. They need a sufficient amount of data to do the stress testing to avoid such loss. As another example, scientists who work in the data science area are often faced with the problem of insufficient data when they are trying to train new and robust machine models.

On the other hand, big data often compromises privacy and results in unjustified analyses because of its immense knowledge. Some European governments implemented the European General Data Protection Regulation in order to prevent misuse of data and violations of privacy rights and to implement strict rules with respect to data protection in order to prevent privacy leaks. This poses a new challenge for the industries that are driven by big data to find solutions that will allow them to make big discoveries while respecting the privacy rights of individuals as well as mandatory government regulations.

One emerging solution is to rely on synthetic data rather than real data, which is statistically very close to real data and can satisfy privacy requirements due to its synthetic nature. However, there are some challenges for tabular data synthesizing tasks. 

The major issues can be summarized as follows: 
\begin{enumerate}
 \item Data Shortage Issue: Generative models aim to produce sufficient outputs with a wide variety. Due to the limited number of input data, it is hard to consider both quantity and variety for synthetic data. Some models may suffer from mode collapse problems, which means the model can not generate various data. Thus it keeps generating the same output.
  \item Data Privacy Issue: Tabular data usually contains users' sensitive information, which could be used to identify individuals and harm their privacy. The more similarity between synthetic data and real data indicates the better quality of synthetic data. However, synthetic data with high similarity could reveal users' information. Therefore, the tabular data synthesizer should try to protect users' information and maintain the high generating quality.
 \item Data Quality Issue (Machine Learning Utility): Synthetic data are used for downstream machine learning model training needs in specific applications. That requires high-quality synthetic data. These high-quality data should satisfy the machine learning utility to complete the downstream tasks. That means if we train the machine learning model using the synthetic data, that should have a similar performance to the machine learning model trained by real data.
  \item Data Quality Issue (Database Constraint):
  Databases are commonly used to store and manage tabular data. Some database constraints define specific properties that data in a database must comply with. The origin data define those properties. The synthetic data with high quality should fulfil these constraints as well.
\end{enumerate}

\begin{figure}[htbp]
    \centering
    \includegraphics[width=\textwidth]{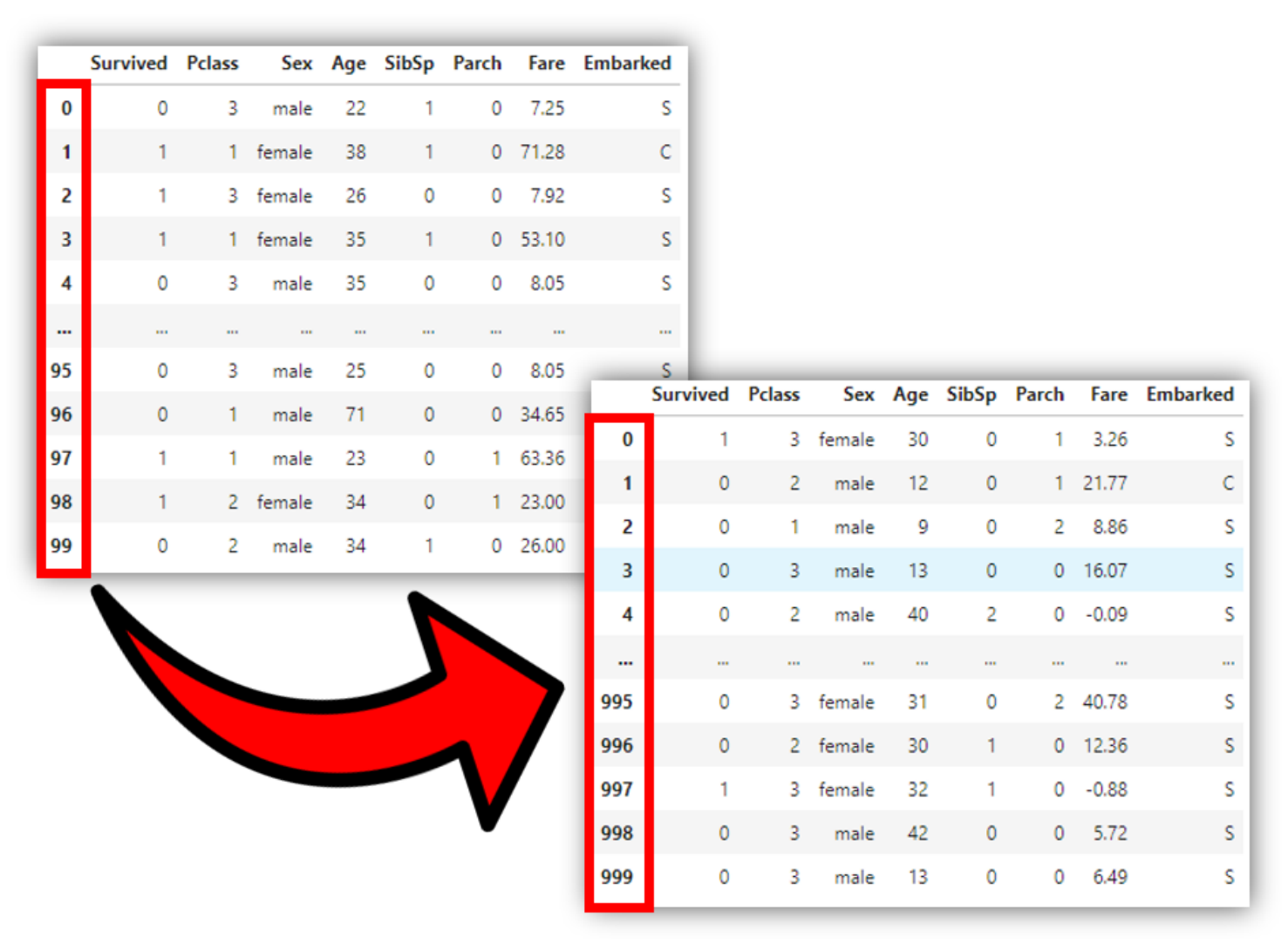}
    \caption{Example of using GAN to solve the tabular data shortage problem}
    \label{fig:GAN_example}
\end{figure}

Generative Adversarial Network (GAN)~\cite{goodfellow} is one of the most promising approaches to synthesize data. Initially, GAN was designed in the past to generate images. Now, it has been migrated to produce tabular datasets as well~~\cite{tgan}~\cite{ctgan}~\cite{octgan}~\cite{tablegan}. Normally, a GAN model is trained on a real dataset. Once it is constructed, it can be used to efficiently generate tabular data. 

Figure \ref{fig:GAN_example} shows the example of using GAN model to solves the \textbf{issue 1} successfully. Most of the recent works pay considerable attention to the \textbf{issue 2} and \textbf{issue 3}. \texttt{TGAN}~\cite{tgan} uses a reversible data transformer and Gaussian Mixture Model (GMM) to pre-process the categorical data, and numerical data further improves the ability to generate categorical data and the distribution of numerical data. The state-of-the-art \texttt{CTGAN}~\cite{ctgan} augments the training procedure with mode-specific
normalization and treats categorical variables as condition vectors and addresses data imbalance by employing a conditional generator. 
\texttt{TableGAN-MCA}~\cite{TableGAN-MCA} proposes a novel Membership Collision Attack against GANs, which allows an adversary given only synthetic entries randomly sampled from a black-box generator to recover partial GAN training data to immune Membership Inference attack. \texttt{ITS-GAN}~\cite{itsgan} studies an incomplete table synthesis problem (using a very small proportion of real data to train GAN models) for tabular data augmentation. It used pre-trained functional dependencies models to enhance the GAN model in order to ensure the generated data satisfied the machine learning utility. \\
Although current works yielding success on the first three issues, \textbf{issue 4} has not been well developed and solved. 
\section{Problem}

Recall the E-commerce problem: the online platforms need to conduct the stress test and use sufficient data to estimate the computational resources required during major shopping events. Required computational resources can be formulated as the query execution cost. The query cost occurs when we take any actions on a table from our database. The query execution cost is computed by combining the cost of each of the operators appearing in the query plan. The standard operators include selection, projection, joint and so on. However, calculating the actual cost of query plans is usually impossible without actually executing the plan. The only method we could calculate the execution cost is to estimate the cost of each operator separately and combine them.
\\
Equation \ref{eqn:query_plan_cost} shows the idea of how to estimate the query execution cost.

\begin{equation}
\label{eqn:query_plan_cost}
\hat{\mathcal{C}}_{\text{Query execution}} = \hat{\mathcal{C}}_{\text{Selection }} + \hat{\mathcal{C}}_{\text{Projection }} + \hat{\mathcal{C}}_{\text{Joint }} + \cdot\cdot\cdot
\end{equation}
where $\hat{C}$ indicates the estimated cost.\\
Despite the current GAN models having made great successes, the existing methods could not ensure their synthetic data fit the requirement for E-commerce platforms as there is a research gap between the \textbf{Issue 4} and current methods. Motivated by this problem, we start from the selection cost and take a further step to selectivity. To maximize the accuracy of selectivity cost, we should let the generated data satisfy the selectivity constraints from the original data.\\
The problem can be stated as:

`How to develop a tabular data generation GAN to model selectivity constraints in tabular data synthesizing’

Motivated by this question, this thesis conducts a series of studies on query selectivity constraint modeling for GAN-based tabular data generation.

\section{Contributions and Thesis Outline}

In this thesis, we design a GAN based tabular data synthesizer that fulfills query selectivity constraints. We propose a novel method to combine with state-of-the-art GAN models by introducing a pre-trained selectivity estimation deep neural network to provide additional control of the selectivity of generated data. By modifying the loss term of the GAN model to ensure the generated data could fit the selectivity constraint. 
We combined the method with two GAN models and tested on five widely used machine learning datasets against three GAN-based tabular data generation methods and one VAE-based method. 

The main contributions of our work can be summarized as following:

\begin{itemize}
    \item Improves the current data reversible transforming method to ensure the GAN model is suitable with any mixed-type data.
    \item Pre-trains a selectivity estimation model. Incorporates the selectivity score in training the generator, thus the synthesizing data can fulfill the selectivity constraints.
    \item The proposed augmentation method is flexible that could compatible any GAN based tabular data synthesizer.
\end{itemize}
The rest of the chapters are organized below:\\
In \cref{chap: Related Work}, we will discuss common approaches of tabular data synthesizing methods and 
the selectivity estimation methods.\\ 
In \cref{chap: Methodology}, we will introduce pre-trained selectivity model, GAN model architecture and how to combine them together.\\ 
In \cref{chap: Experiments}, we will talk about the implementation, parameter setting and present sufficient experimental results. 
\\
In \cref{chap: Conclusion and Future Work}, we will gave a summary of the experiments, proposes the limitation of our method, and provides the future improvement direction.

\clearpage   

\lhead{\emph{Chapter 2 Related Work}}

\chapter{Related Work}

\label{chap: Related Work}

In this chapter, we will introduce some background information regarding traditional generative models for tabular data generation, Bayesian networks in statistics, variational autoencoders (VAEs) and generative adversarial networks (GANs) in computer science. Furthermore, we will discuss the different approaches to selectivity estimation, which include traditional estimation models and novel regression-based estimation models.
\section{Tabular Data Generative models}

Generative models are unsupervised machine learning models that attempt to discover the regularities, patterns and distributions from the input origin data, then use the learned knowledge to generate plausible data. The purpose of synthetic data generation is to resolve four issues, which are discussed in \cref{chap: Introduction}: data shortage issues, data privacy issues, data quality issues, including Machine Learning and Data Base issues. The models learn from the existing real data and generate their distributions from the acquired data, fulfilling requirements from the different industries and addressing the four issues.

Statistical generative models such as Bayesian networks and Gaussian mixture models are suitable for fitting certain probability distributions. However, in the real world, the datasets are often more complex and come in different formats. As a result, the statistical models are not usually compatible with image or text datasets.
\subsection{Bayesian network}

The Bayesian network~\cite{BN} model is widely studied in the field of statistical and machine learning. 

\begin{figure}[htp]
    \centering
    \includegraphics[width=10cm]{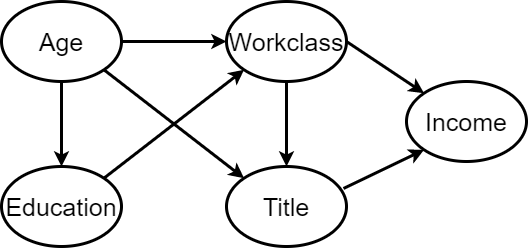}
    \caption{Bayesian network over five attributes from \texttt{PrivBayes}~\cite{Privbayes}}
    \label{fig:PrivBayes}
\end{figure}

Assume $A$ is the set of attributes on the dataset $D$. $D$ has a joint probability distribution over the cross-product of $A$’s attribute domains. A Bayesian network can be used to describe the distribution through the particular conditional independence between the attributes of $A$. To be specific, a Bayesian network is a directed acyclic graph (DAG) that represents each attribute in $A$ as a node and conditional independence between attributes using directed edges. A simple Bayesian network schematic is shown in \fref{fig:PrivBayes}. Bayesian networks are simple but powerful graphical models. They can approximate the complete-dimensional data distribution by combining low-dimensional data distributions.

The standard Bayesian network generating synthetic data can be summarized as following steps:
\begin{enumerate}
    \item Train a standard Bayesian network
    \item Compute the differential privacy distributions of the data and then inject the Laplace noise to each parameter of the learned Bayesian network
    \item Generate synthetic data from the noisy Bayesian network
\end{enumerate}
However, the inappropriate amount or content of noise would lead the Bayesian network to a very poor generation performance. The other limitation of the traditional Bayesian networks is that they cannot handle continuous data, but they can represent a joint distribution of discrete variables.
\clearpage
\subsection*{PrivBayes}

\texttt{PrivBayes}~\cite{Privbayes} was developed by ZHANG et al. in 2017. A novel Bayesian Network-based model provides a solution to protect differential privacy. The formal definition of $\varepsilon$-differential privacy is as follows:\\
$$
\Pr[G(D_1) = O] \leq e^{\varepsilon} \cdot \Pr [G(D_2) = O]
$$
where $\Pr[\,\cdot\,]$ is the probability of an event.\\
\texttt{PrivBayes} uses traditional Bayesian network architecture but a differential privacy learning algorithm to reduce the amount of noise that needs to be inserted. They compute a differential private Bayesian network that approximates the full-dimensional distribution using the Laplace mechanism and the exponential mechanism. Before the model, it discretized all continuous variables into 16 equal-sized bins to make the model more flexible for mixed-type datasets.\\\\
The \texttt{PrivBayes} can be constructed in to three stages:
\begin{enumerate}
    \item Using $\varepsilon_1$-differential privacy method to build a $k$-degree Bayesian network $N$ through the attributes in dataset $D$.
    \item Using $\varepsilon_2$-differential privacy method to generate $d$, a set of conditional distributions for $D$.
    For example, each pair of conditional distribution $\Pr[X_i|\Pi_i]$ got a noisy distribution $\Pr^{\star}[X_i|\Pi_i]$.
    \item Use the Bayesian network $N$ and the $d$ noisy conditional distributions to derive an estimated distribution of the tuples in $D$, then  sample tuples from the estimated distribution to generate a synthetic dataset $D^{\star}$.
\end{enumerate}

In phase 1, the choice of $k$ is non-trivial. It involves a trade-off between the Bayesian network’s original quality. A Bayesian network with a larger k keeps more information from the full-dimensional distribution Pr[A]. For instance, a $(d - 1)$ - degree Bayesian network can fit the distribution. In contrast, if a 1 - degree Bayesian network is present, there is much information loss when fitting the distribution. To resolve this problem, they use a measure called $\theta$ - usefulness to provide a more choose k automatically to balance the accuracy of the Bayesian network. Through the constraint of $\theta$ - usefulness, \texttt{PrivBayes} uses a greedy algorithm to maximize the mutual information and to optimally structure the tree-based Bayesian network.

\subsection{Autoencoder}
\begin{figure}[htp]
    \centering
    \includegraphics[width=15cm]{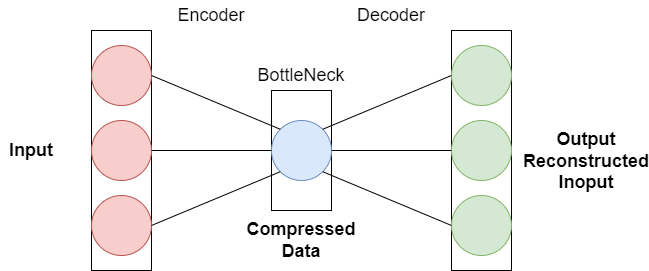}
    \caption{Autoencoder scheme from ~\cite{bottleneck}}
    \label{fig:AE_bottle}
\end{figure}

Back in 1987, Autoencoder~\cite{autoencoder} was first developed by Ballard.
An autoencoder is a type of deep neural network used to solve an unsupervised learning task --- representation learning. That means the autoencoder could efficiently learn the coding of datasets. Thus, it can usually be used to remove the redundancy and extract the important data features. More specifically, a deep neural network contains a bottleneck inside the network and then forces a compressed knowledge representation from the input data. \fref{fig:AE_bottle} shows the sample scheme for autoencoder.

A standard autoencoder contains two components: an encoder $\mathcal{E}(\,\cdot\,)$ and a decoder  $\mathcal{D}(\,\cdot\,)$. The encoder compresses the input and produces the code, the decoder then reconstructs the input only using this code.\\
To be more specific, provide a dataset $\textbf{X}$,  the encoder $\mathcal{E}(\,\cdot\,)$ compresses the input data $x$ from $\textbf{X}$  into a hidden distributed representation $z$ (Compressed data from \fref{fig:AE_bottle}). The encoding step can be shown as:
$$z = \mathcal{E}(x)\text{, \,where } x \sim \textbf{X}.$$
Then the decoder $\mathcal{D}(\,\cdot\,)$ takes the hidden representation $z$ and reconstructs it. Lastly, it will produce $\hat{x}$:
$$
\hat{x} = \mathcal{D}(z)\text{, \,where } \hat{x} \approx x.
$$
The generated data $\hat{x}$ is approximate to the original data $x$ because a successful autoencoder aims to extract all the essential features of $x$. The $x$ and $\hat{x}$ should share the same properties. Therefore, the autoencoder network is trained by minimizing the reconstruction error:
$$
\mathcal{L} = (x,\hat{x}) = \mathbb{E}_{x\sim\textbf{X}}[\| \mathcal{D}(\mathcal{E}(x)) - x \|^2_2]
$$
Typically, the reconstruction error is the mean squared error, which measures the differences between the original input data and the later reconstruction $\hat{x}$. 
\subsection*{Variational Autoencoder}
The Variational Autoencoder (VAE)~\cite{VAE} is a variant of standard autoencoder. The~`variational' means that the encodings distribution is regularised during the training process to keep the approximate features and generate new data.\\ The architecture of a VAE is the same as a standard autoencoder. It contains an encoder $\mathcal{E[\,\cdot\,]}$ and a decoder $\mathcal{D[\,\cdot\,]}$. The VAE is trained using the similar reconstruction error $\mathcal{L} = (x,\hat{x})$, which aims to minimize the difference between origin data and the generated data as well. \\As mentioned before, some regularisation term is introduced to VAE. A small modification is applied to the encoding-decoding step to achieve the regularisation. In a traditional autoencoder, the input data is a single point $x$ from the original data $\textbf{X}$, and the output is the hidden representation $z$. In VAE, the hidden representation is seen as a Gaussian distribution $\mathcal{N}[\,\cdot\,]$. Therefore the hidden representation from a VAE encoder forms a normal distribution $\mathcal{N}(\mu, \sigma^2)$.
$$
\mu, \sigma = \mathcal{E}(x)\text{, \,where } x \sim \textbf{X}.
$$
The decoder takes one sample $z$ from the $\mathcal{N}(\mu, \sigma^2)$ and reconstructs the data.
$$
\hat{x}= \mathbb{E}_{z\sim\mathcal{N}(\mu,\sigma)}[\mathcal{D}(z)]\text{, \,where } \hat{x} \approx x.
$$
Additionally, VAE made a constraints to force all the aggregated distribution of $z$ over all  the data $\textbf{X}$ to be $\mathcal{N}(0,\textbf{I})$. Under this constraint, we can input any vector sampled from $\mathcal{N}(0,\textbf{I})$ into the trained decoder to generate new data.\\ The encoder-decoder pair is multi-input and multi-output deep neural networks and trained by stochastic gradient descent (SGD). The \eref{eqn:VAE_loss} shows the reconstruction error is further modified as a evidence lower-bound (ELBO) loss.

\begin{equation}
\label{eqn:VAE_loss}
\mathcal{L} = [\,\,\,\mathbb{E}_{x\sim\mathcal{N}(\mu,\sigma\textbf{I})}[\| \mathcal{D}(\mathcal{E}(x)) - x \|^2_2]\,\,\, + \,\,\, \mathbb{KL}(\mathcal{N}(\mu,\sigma\textbf{I})\|\mathcal{N}(0,\textbf{I}))\,\,\,].
\end{equation}
The first term is precisely the same as the traditional autoencoder. The second term $\mathbb{KL}(p||q)$ is the Kullback–Leibler (KL) divergence. The KL divergence is used to measure the distance between two distributions $p$ and $q$ using the following formula:
$$
\mathbb{KL}(p||q) = - \int_x p(x)\,\,\log\,\frac{q(x)}{p(x)}
$$
In the VAE case, we have the $p$ as the standard normal distribution $\mathcal{N}(0,\textbf{I})$, the $q$ as the distribution of $z \sim \mathcal{N}(\mu,\sigma\textbf{I})$. Thus, this term makes a constraint to ensure the $z$ to be the standard normal distribution. The outer expectation is computed by taking the average over minibatch. The training process will end when the model converges. The learned $\mathcal{D}$ is an approximated mapping from a multivariate Gaussian distribution to the data distribution. \\

\subsection{Generative Adversarial Network}
Generative Adversarial Networks (GANs) were firstly proposed by Goodfellow~\cite{goodfellow} in 2014. GAN is a generative model using deep learning techniques to generate different types of data to fit the requirements of variant industry needs.

Generative modelling is an unsupervised learning task.
GAN models convert the problem from an unsupervised learning task to a supervised learning task masterly using two sub-models: A generator $\mathcal{G}$ and a discriminator $\mathcal{D}$. 
\begin{figure}[htp]
    \centering
    \includegraphics[width=13cm]{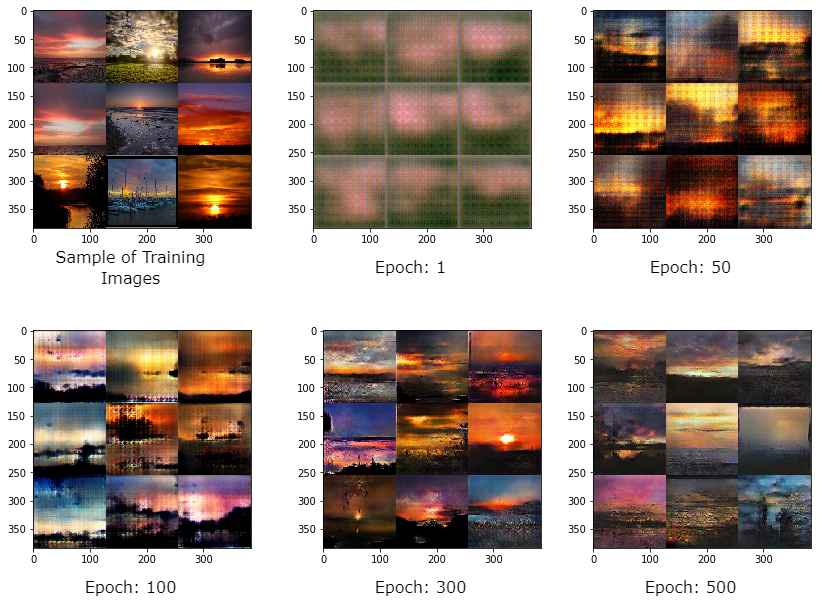}
    \caption{Training evolution of \texttt{DCGAN}~\cite{DCGAN}}
    \label{fig:DCGAN}
\end{figure}

The generator $\mathcal{G}$ is trained to generate new samples, and the discriminator $\mathcal{D}$ is required to recognize if the generated samples are real or fake. The training process is like an adversarial zero-sum competition as the $\mathcal{G}$ have to foolish the $\mathcal{D}$ and the $\mathcal{D}$ tries its best to classify the provided data. During the training, the $\mathcal{G}$ and $\mathcal{D}$ grow together, the quality of generated data is higher and higher, as well as the ability to recognize $\mathcal{D}$. \fref{fig:DCGAN} shows the training evolution of a deep convolutional generative adversarial network~\cite{DCGAN} which is commonly used to generate images. It is an example of generating sunset images from the first epoch to the 500th epoch. The growth of $\mathcal{G}$ and $\mathcal{D}$ shows the adversarial process during training.

\subsection*{Vanilla GAN}

\begin{figure}[htp]
    \centering
    \includegraphics[width=13cm]{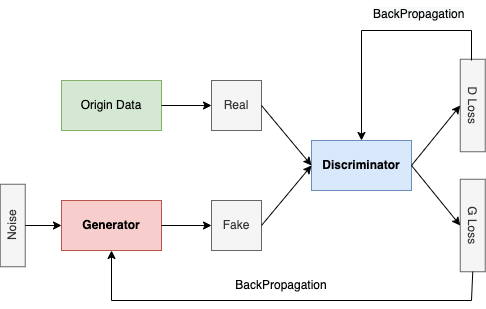}
    \caption{GAN Training Process}
    \label{fig:trainGAN}
\end{figure}

As mentioned in the last section, the GAN model contains two deep neural networks that make the training process quite complete; it has to solve those complications:
\begin{itemize}
    \item Handle two different training tasks (Generating and Classification)
    \item Identify training convergence
\end{itemize}
\fref{fig:trainGAN} shows the flow chart for Vanilla GAN training. Vanilla GAN is the origin GAN training method proposed by Goodfellow~\cite{goodfellow}. The main training process can be broken down into two alternating steps:

\begin{enumerate}
    \item Update Discriminator $\mathcal{D}$.
    \item Update Generator $\mathcal{G}$.
\end{enumerate}
The two steps run repeatedly to continuous training the $\mathcal{G}$ and $\mathcal{D}$.

\textbf{Step 1:}\\
The Discriminator $\mathcal{D}$ is a classifier which needs to distinguish if the data is generated by the Generator $\mathcal{G}$. To train the classifier more accurately, we should use the data from both data sources. The real data from the original data as the positive samples. The fake data generated by $\mathcal{G}$ as the negative samples. During the training, the $\mathcal{D}$ classifies both real data and fake data and produces a D loss. The discriminator loss will penalize the $\mathcal{D}$ for all misclassified samples. The $\mathcal{D}$ updates its weights through backpropagation from the D loss function \eref{eqn:DLoss}:

\begin{equation}
\label{eqn:DLoss}
 \mathcal{L_\mathcal{D}} = - \mathbb{E}_{x\sim\textbf{X}}[\log(\mathcal{D}(x))] + \mathbb{E}_{z\sim\mathcal{N}(0,\textbf{I})}[\log(1-\mathcal{D}(\mathcal{G}(z)))]   
\end{equation}
$\mathcal{L_\mathcal{D}}$ is the cross-entropy loss for binary classification.
\\\\
\textbf{Step 2:}\\
The Generator $\mathcal{G}$ is used to create fake data incorporating feedback from the Discriminator. It aims to fool the $\mathcal{D}$ and let $\mathcal{D}$ can not complete the classification task well. The training process for $\mathcal{G}$ is more complicated. The whole training process involves three components: Random Input which makes sure the GAN produce a wide variety of data; Generator $\mathcal{G}$ which generates the data using the random feed input and the D loss, which enhances the generating ability by penalizes the $\mathcal{G}$ for correctly classified samples. The Generator $\mathcal{G}$ is optimized using the equation:
\begin{equation}
\label{eqn:GLoss}
 \mathcal{L_\mathcal{G}} = \mathbb{E}_{z\sim\mathcal{N}(0,\textbf{I})}[\log(\mathcal{D}(\mathcal{G}(z)))]   
\end{equation}
\\\\
\textbf{Minimax Loss Function:}\\
The MiniMax loss function(\eref{eqn:minimax}) is commonly uses in standard GAN which is combined from the $\mathcal{L_\mathcal{G}}$(\eref{eqn:GLoss}) and $\mathcal{L_\mathcal{D}}$(\eref{eqn:DLoss}). In this function the generator tries to minimize the following function while the discriminator tries to maximize it.
\begin{equation}
\label{eqn:minimax}
\mathbb{E}_{x\sim\textbf{X}}[\log(\mathcal{D}(x))] + \mathbb{E}_{z\sim\mathcal{N}(0,\textbf{I})}[\log(1-\mathcal{D}(\mathcal{G}(z)))]   
\end{equation}

In the first term, $\mathcal{D}(x)$ is the probability estimation to indicate if the real instance $x$ is real. The $E_x$ is the expected value among all real data instances. In the second term, $\mathcal{G}(z)$ is the generated data using the input noise $z$. $\mathcal{D}(\mathcal{G}(z))$ is the probability estimation to indicate if fake instance $\mathcal{G}(z)$ is real. The $E_z$ is the expected value among all fake data instances. The formula derives from the cross-entropy between the real and fake distributions. 
However, the Vanilla GAN always suffers from the mode collapse problem. Mode collapse means the Generator keeps producing the same output data. 

\subsection*{Wassersttein GAN}
Wassersttein GAN ~\cite{WGAN} is a improved training method to solve the mode collapse problem. WGAN uses a critic network $\mathcal{C}[\,\cdot\,]$. The output of the critic network $\mathcal{C}[\,\cdot\,]$ is dynamic. When the input is more realistic, the output value is larger or vice versa.\\
To achieve this, the critic network is trained using:

\begin{equation}
\label{eqn:critic}
 \mathcal{L_\mathcal{C}} = - \mathbb{E}_{x\sim\textbf{X}}[\log(\mathcal{C}(x))] + \mathbb{E}_{z\sim\mathcal{N}(0,\textbf{I})}[\log(1-\mathcal{C}(\mathcal{G}(z)))]   
\end{equation}

That means, the critic network is trying to maximize the output on real data and minimized the output of the generated data. Conversely, the generator is trying to maximize the output of the critic network using the following function:

$$
 \mathcal{L_\mathcal{G}} = - \mathbb{E}_{z\sim\mathcal{N}(0,\textbf{I})}[\log(1-\mathcal{C}(\mathcal{G}(z)))]   
$$
WGAN aims to minimize the Wasserstein distance between the generated and real data distribution. The  Wasserstein distance will provide the between probability distributions on a given metric space. The Wasserstein distance $\mathbb{W}(\,\cdot\,)$can be shown as:
$$
\mathbb{W}(\textbf{x},\mathbb{P}_\mathcal{G}) = \sup_{\|f\|_L\le1} \mathbb{E}_{x\sim\textbf{X}}[f(x)] - \mathbb{E}_{\hat{x}\sim\mathbb{P}_\mathcal{G}}[f(\hat{x}]
$$
where $\|f\|_L\le1$ indicates $f$ is a 1-Lipschitz function.\\
\eref{eqn:critic} can be seen as approaching the Wasserstein distance equation. Therefore, the training process for the Generator can be seen as minimizing the Wasserstein distance. Note that parameters in the $\mathcal{C}$ are controlled to fit the  1-Lipschitz constraint.\\\\
The WGAN is also trained by stochastic gradient descent (SGD) with the similar steps to Vanilla GAN: 
In the first step, WGAN updates Critic $\mathcal{C}$, the next step is to update Generator $\mathcal{G}$. The two steps repeatedly run to continuous training the $\mathcal{C}$ and $\mathcal{D}$.\\
 Mostly, the training processes are the same as the Vanilla GAN training processes. Besides, we use Critic $\mathcal{C}$ instead of Discriminator $\mathcal{G}$. Additionally, the parameter in $\mathcal{C}$ has to be modified to satisfy the 1-Lipschitz condition in step 1.

\subsection{GAN Variants for Tabular Data Generation}
\label{sec:GAN Variants for Tabular Data Generation}
The initial GAN model was used to synthesize the image~\cite{goodfellow}. The development of GAN has led to more and more varieties being proposed in different fields. Now, GAN models have migrated from image data to tabular data. This session summarizes the most commonly used GAN-based tabular data generation models.

\begin{table}[!h]
    \centering
    \begin{tabular}{c|c|c|c}
    Model  & Discrete & Numerical & Enhanced\\\hline\hline
       \texttt{MedGAN} (2017)  & &\checkmark & High-Dimensional\\
       \texttt{tablgeGAN} (2017) & & \checkmark & Data Semantic\\
       \texttt{CTGAN} (2019) &\checkmark& \checkmark & Imbalanced Data\\
        \texttt{OCTGAN} (2021) &\checkmark& \checkmark & Imbalanced Data\\
    \end{tabular}
    \caption{Summary for commonly used Tabular data generation GAN models}
    \label{tab:ganmodels}
\end{table}
\subsubsection*{MedGAN}
\begin{figure}[htp]
    \centering
    \includegraphics[width=6cm]{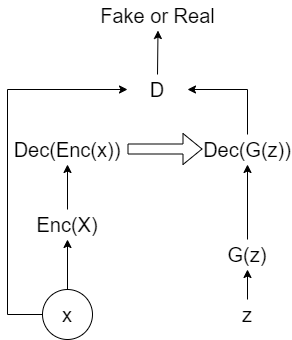}
    \caption{Architecture of \texttt{MedGAN}}
    \label{fig:Medgan}
\end{figure}

 \texttt{MedGAN}~\cite{Medgan} was proposed by Choi et al. in 2017 to produce high-dimensional electronic health record (EHR) data in discrete variables. High-dimensional data suffers from the curse of dimensionality. To overcome this, \texttt{MedGAN} develop an autoencoder to learn the data representation. The original data can be represented as a low-dimensional data representation without any information loss by using the autoencoder.\\
 \texttt{MedGAN} is limited applied to binary responses and continuous features. The binary features are represented as 1 or 0. The continuous features should apply a min-max normalization method to normalized to the range [0,1] using the MinMax normalization formula:
 
 \begin{equation}
 \label{eqn:minmaxnormal}
      \frac{c_{i,j}-min(C_i)}{max(C_i)-min(C_i))}
 \end{equation}

 where $C_i$ means the ith continuous column and $c_{ij}$ means the jth value in the ith continuous column.\\
 \fref{fig:Medgan} shows the Architecture of \texttt{MedGAN}. The discrete $x$ comes from the source EHR data, $z$ is the random prior for the generator $\mathcal{G}$; $\mathcal{G}$ is a feedforward network with shortcut connections (right-hand side figure); An autoencoder (i.e, the encoder \textbf{Enc} and decoder \textbf{Dec}) is learned from $x$; The same decoder
 \textbf{Dec} is used after the generator $\mathcal{G}$ to construct the discrete output. The discriminator $\mathcal{D}$ tries to differentiate real input $x$ and discrete synthetic output $\textbf{Dec}(\mathcal{G}(z))$. The pre-trained autoencoder is used to learn discrete features, and then it can be applied
to decode the continuous output of G. The autoencoder uses to mean a squared loss for the loss function to check if only continuous features exist in the generated data and cross-entropy loss to check if the discrete columns are in binary. The loss function for the GAN model is the standard Minimax loss function (see \eref{eqn:minmaxnormal}) from Vanilla GAN~\cite{goodfellow}.
The loss function for the autoencoder is the mean squared error if the table contains
only continuous columns and cross-entropy loss if the columns are all binary. The
generator and discriminator are trained using the same loss function as a vanilla
GAN.\\However, the \texttt{MedGAN} does not support tabular data with mixed data types. Only continuous and binary discrete data is acceptable. The real-world data is usually complicated with the mixed data type, and this is not very suitable in most real-world scenarios. 
\subsubsection*{Table-GAN}
\texttt{Table-GAN}~\cite{tablegan} is a variation on the GAN Architecture published by Park et al. in 2017. It applies the idea of \texttt{DCGAN}~\cite{DCGAN} which is a typically used image generation GAN model to generate tabular data to protect people‘s privacy. In this paper, \texttt{Table-GAN}  is developed against three attacks: re-identification attack, attribute disclosure and membership attack.

\begin{figure}[htp]
    \centering
    \includegraphics[width=15cm]{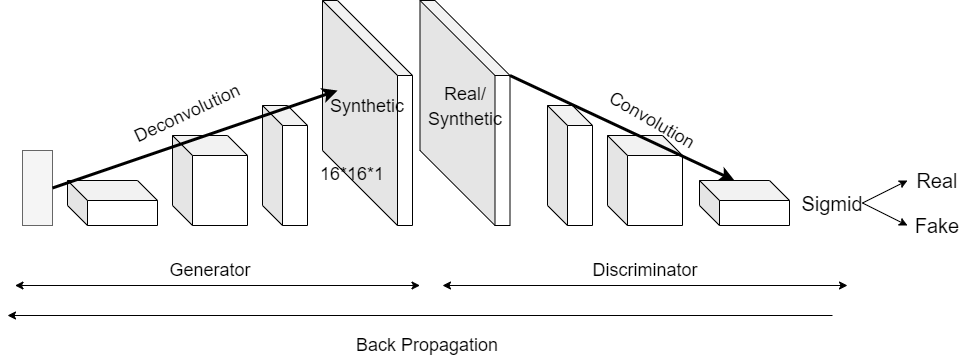}
    \caption{Architecture of \texttt{table-GAN}}
    \label{fig:tablegan}
\end{figure}

The \texttt{Table-GAN} is developed from Deep Convolutional Generative Adversarial Network. Unlike the GAN model mentioned in the previous section, the Generator $\mathcal{G}$ and the Discriminator $\mathcal{D}$ are a pair of Convolutional neural networks and De-convolutional neural work. Additionally, \texttt{Table-GAN} add another neural network Classifier $\mathcal{C}$ to supervise the semantic.
Since  \texttt{DCGAN}~\cite{DCGAN} was used for image generation. Thus the input data is not a vector but a matrix with a number. Thus, all the data should be prepossessed before training. For continuous variables, the MinMax normalization \eref{eqn:minimax} is applied; the discrete variables are converted to floating-point numbers or one-hot vectors. After that, the preprocessed data should be re-arranged into a squared matrix. If the number of columns can not fill in a squared matrix, zeros are padded behind to increase the vector length to form a squared matrix. For example, if the vector length for preprocessed features is 12, then four zeros are padding behind, then the vector after padding can form a $4 \times 4 $ matrix.
\\\\
\fref{fig:tablegan} shows the architecture of the Generator $\mathcal{G}$ and the Discriminator $\mathcal{D}$  from \texttt{Table-GAN}. The Generator $\mathcal{G}$ performs a series of deconvolution operations to generate data, while the Discriminator $\mathcal{D}$ has the corresponding convolution layers to classify the real and fake data. The final loss after the sigmoid activation can be back-propagated to the Generator. The dimensions of the latent vector input $z$ and intermediate tensors should be configured considering the number of attributes (e.g.,
16 × 16 = 196 attributes in this figure). The Classifier $\mathcal{C}$ increases the semantic integrity of synthetic
records that have the same structure of the Discriminator $\mathcal{D}$. For example, (cholesterol=50, diabetes=1) is not a correct record because cholesterol=50 is too
low to be diagnosed as diabetes. The $\mathcal{C}$ is trained by the ground-truth label from the original table, therefore it can recognize if the generated data is semantic correct. 
 \\
The \texttt{Table-GAN} is trained using standard GAN loss function (\eref{eqn:minimax}) with two additional term.\\
\textbf{Information Loss}\\
The information loss is defined as the discrepancy between
two statistics of synthetic and real records. The information loss compares the first-order statistics(mean) and second-order statistics(sd) using the \eref{eqn:tablegan_meansd}:\\
\begin{equation}
\label{eqn:tablegan_meansd}
  \begin{aligned}
    \mathcal{L}_{mean} &= \|\mathbb{E}[\textbf{f}_x]_{x\sim p_{data}(x)} - \mathbb{E}[\textbf{f}_{\mathcal{G}_{(z)}}]_{z\sim p(z)}\|_2\\
        \mathcal{L}_{sd} &= \|\mathbb{S}\mathbb{D}[\textbf{f}_x]_{x\sim p_{data}(x)} - \mathbb{S}\mathbb{D}[\textbf{f}_{\mathcal{G}_{(z)}}]_{z\sim p(z)}\|_2
      \end{aligned}
\end{equation}
The less value of $\mathcal{L}_{mean}$ and $\mathcal{L}_{sd}$  indicates that real and synthetic records have the statistically same features from the perspective of the discriminator. To make the privacy degree more controllable, the two loss terms are combined with two thresholds using \eref{eqn:tablegan_information}.
\begin{equation}
\label{eqn:tablegan_information}
    \mathcal{L}_{info}^G = \max(0,\mathcal{L}_{mean}-\delta_{mean}) + \max(0,\mathcal{L}_{sd}-\delta_{sd})
\end{equation}

\textbf{Classification Loss}\\
Classification loss maintains the semantic integrity using the \eref{eqn:tablegan_cl}. It measures the
between the label of a generated record and the label predicted by the classifier
for that record.

  \begin{equation}\label{eqn:tablegan_cl}
  \begin{aligned}
    \mathcal{L}_{class}^{\mathcal{C}} &= \mathbb{E}[|l(x)-\mathcal{C}(remove(x))|]_{x\sim p_{data}(x)},\\
        \mathcal{L}_{class}^{\mathcal{G}} &= \mathbb{E}[|l(\mathcal{G}(x))-\mathcal{C}(remove(\mathcal{C}(x)))|]_{z\sim p(x)}
 \end{aligned}
\end{equation}
where $l(\,\cdot\,)$ is a function that returns the ground truth label, $remove(\,\cdot\,)$ is to remove the label attribute and $\mathcal{C}(\,\cdot\,)$ is a label predicted by the classifier neural network.\\
\texttt{Table-GAN} is a novel approach for tabular data generation with convolutional neural network, but the CNN and classifier architecture restrict it only compatible with limited data type.

\subsubsection*{CT-GAN}
\label{sec:ctgan}
\begin{figure}[htp]
    \centering
    \includegraphics[width=8cm]{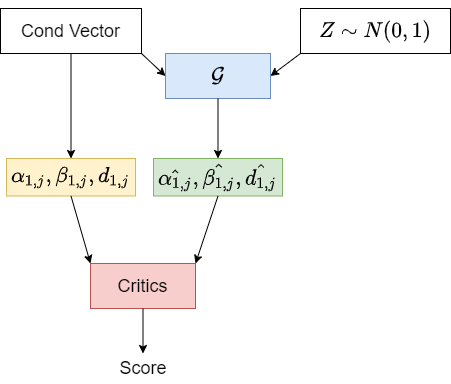}
    \caption{Architecture of \texttt{CTGAN}}
    \label{fig:ctgan}
\end{figure}

\texttt{CTGAN}~\cite{ctgan} was proposed by Xu et al. from MIT in 2019. This paper uses Mode-specific normalization to solve mixed data type and Non-Gaussian distribution assumption problems and uses Conditional vectors to solve imbalanced categorical column problems. \\
\textbf{Data Preprocessing}\\
The discrete columns are represented in a one-hot vector, the ith discrete column is donated as $d_i$. A special method called Mode-specific normalization is used to process continuous values. This method is used since the continuous columns in the real-world are usually not following a normal distribution but a Multi-modal distribution. In this paper, for each continuous column $C_i$, variational Gaussian mixture model is used to find the number of Modes. Then fit a Gaussian mixture and find the parameters Mean, Weight and Standard Deviation of a mode respectively. For each value $c_{i,j}$ in $C_i$ compute the probability of $c_{i,j}$ coming from each mode. Using the most possible mode to normalize $c_{i,j}$, the normalized value donates as  $\alpha_{i,j}$ and the chosen mode is represent in one-hot vector $\beta_{i,j}$. \\
After that, the representation of a row becomes the concatenation of continuous and discrete columns can be written as follows:
$$
r_j = \alpha_{1,j}\oplus \beta_{i,j} \oplus ...\alpha_{N_c,j}\oplus \beta_{N_c,j} \oplus d_{1,j}...\oplus d_{N_d,j}
$$
\textbf{Conditional Generator}\\
Usually, when a GAN model is trained using an unbalanced dataset, the data in the minor category will not be sufficiently represented. \texttt{CTGAN} uses a conditional generator to enforce that the Generator matches a given category.\\
The \textbf{cond vector} is introduced to indicate the condition. Recall that after the data preprocessing, all the discrete columns $D_1 ....D_N$ end up as one-hot vector $d_1 ...d_{Nd}$, such that the ith one hot vector is $d_i$. The one-hot representation vector is the mask vector $m_i$ for each discrete column $D_i$. All mask vectors are concated to form a \textbf{cond vector}. For instance, for two discrete
columns, $D_1 = \{1, 2, 3\}$ and $D_2 = \{1, 2\}$, the condition for $D_2$ assigned to 1. Thus, the $D_1$ mask vectors $m_1 = [0, 0, 0]$ since no condition is assign to $D_1$, and the $D_2$ mask vectors  $m_2 = [1, 0]$; The final $\textbf{cond vector} = [0, 0, 0, 1, 0]$.\\\\\texttt{CTGAN} use fully connected hidden layers in both generator and critics (discriminator). For the generator, the input is a random variable z and output is our synthetic data. For the Critics, the PACGAN~\cite{pacgan} framework is used to prevent mode collapse. The model is trained using WGAN loss(\eref{eqn:critic}) with gradient penalty and Adam optimizer. 
\fref{fig:ctgan} shows the architecture of \texttt{CTGAN}, the \textbf{cond} vector feed into the conditional generator, that generates synthetic conditioned rows. With training-by-sampling, the \textbf{cond} and training data are sampled according to the log-frequency of each category, thus \texttt{CTGAN} can evenly explore all possible discrete values.
 \\
Now, \texttt{CTGAN} has become a commonly used GAN-based tabular data generation method, and it can generate data effectively with high-quality synthetic. Thus, \texttt{CTGAN} is one of the models we used as the base model to further develop our method.

\section{Selectivity Estimation}
Query execution cost estimates the number of computational resources (e.g. CPU and I/O resources) for running queries in the current query plan. The total query plan cost is computed by combining the cost of each of the operators appearing in the plan. Selection, Projection and Joint are three commonly used standard operators. In this thesis, we focus on the selection operation. Therefore, this session will introduce some regular approaches to the selectivity estimation method. In database systems, selectivity estimation has been extensively studied. Common approaches are sampling~\cite{sampling} and histograms~\cite{histo}. However, most of them suffer from the curse of dimensionality, which means they could not be compatible with the high-dimensional data. Mattig et al.~\cite{Kernal} use the Kernel-based cardinality to highlight the distribution of metric space. However, the kernel function usually needs to be supported by strong assumptions, and a single kernel function sometimes is insufficient to solve the complicated inner correlation of high-dimensional data.

\subsection{Regression-based Models}
\label{sec:selnet}
Another method is to see the Selectivity estimation as a regression problem with query object and threshold as input features. Wang et al.~\cite{selnet} developed a Regression based model called \texttt{Selnet} to estimate the selectivity as well as maintain the  consistency for high-dimensional data. Definition \ref{def:selectiviy} shows how to form the selectivity to a regression problem.
\begin{definition}[Selectivity]
\label{def:selectiviy}
Given a database with $d$ dimensional vector $\mathbb{D}=\{\textbf{o}_i\}^{n}_{i = 1}$, $\textbf{o}_i \in \mathbb{R}^d$. Provide a distance function $dist(\,\cdot\,)$, a scaler threshold $t$ and a query object $x\in\mathbb{R}^d$. The estimate the selectivity in the database can be written as: $$|\{\,\textbf{o}\,|\,dist(\textbf{x},o)\le t,\textbf{o}\in \mathbb{D}\}|$$
The selectivity (i.e., the ground truth label) $y$ of a query object $\textbf{x} $ and a threshold $t$ as generated by a \textbf{value function}:
$$y = f(\textbf{x},t,\mathbb{D})$$
\end{definition}

Thus, the selective estimation can be seen as estimate $f(\textbf{x},t,\mathbb{D})$ using $\hat{f}(\textbf{x},t,\mathbb{D})$. However, $f$ is complex, thus $f$ is broke into small sub-functions to rather than using one function to estimate $f$ directly. For example, let $y = y_1 + y_2$ and $t = t_1 + t_2$. Then we can use two linear model to estimate corresponding $y_1$ and $y_2$ with $t_1$ and $t_2$ in the range of $[0, t_1]$ and $(t_1, t_2]$. Following this idea, the Threshold Partitioning method (Definition \ref{def:Threshold}) is adopt.
\begin{definition}[Threshold Partitioning]
\label{def:Threshold}
Assume the maximum threshold is $t_{max}$, the $t_{max}$ is divided it with an increasing sequence of $(L+2)$ value: $[\tau_0,\tau_1,\cdot\cdot\cdot,\tau_{L+1}]$. We have $\tau_0 = 0$ and  $\tau_{L+1} = t_{max}+\epsilon$ where $\epsilon$ is a small positive quantity to cover corner cases. Let $g_i(\textbf{x},t)$ be an interpolant function for interval $[\tau_{i-1}, \tau_i)$ and we have:

\begin{equation}
\label{eqn:threshold}
    \hat{f}(\textbf{x},t,\mathcal{D}) = \sum^{L+1}_{i = 1} \textbf{1}[t\in [\tau_{i-1}, \tau_i)] \, \cdot \, g_i(\textbf{x},t)
\end{equation}
Where $\textbf{1}[\,\,\,]$ is the indicator function.
\end{definition}
\textbf{Selnet} uses $L+1$ continuous piece-wise linear functions to implement $g_i(x,t)$ for the range of $[\tau_{i-1}, \tau_i)$ from \eref{eqn:threshold}. The $\tau_i$ values are called \textbf{control points}. Let $p_i$ be the estimated \textbf{Selectivities at Control Points $\tau_i$}. 
The $g_i$ function can written using  $\tau_i\text{ and }p_i$ through:
\begin{equation}
\label{eqn:g}
    g_i(\textbf{x},t) = p_{i-1} + \frac{t-\tau_{i-1}}{\tau_i - \tau_{i-1}} \,\cdot\,(p_i - p_{i-1})
\end{equation}
Hence, $\tau_i\text{ and }p_i$ are the two parameters for each sub-regression model. The final estimation function can be re-parameterized as $\hat{f}(\textbf{x},t,\mathbb{D};\Theta)\text{ where }\Theta\text{ represents } \{(\tau_i,p_i)\}^{L+1}_{i=0}$. Regression-based estimation methods need to find the best control points and estimate the corresponding $p_i$, then combine the $\tau_i\text{ and }p_i$ to estimate the total selectivity $y$.

\clearpage   

\lhead{\emph{Chapter 3 Methodology}}

\chapter{Methodology}

This chapter will discuss the metrics we used and the methods used in our experiments. In section \ref{sec:proposedmethod}, we will discuss the proposed methodology we use to enhance the GAN model to fulfil query selectivity constraints. Section \ref{sec:evaluationmetrics} will introduce the metrics used to measure the quality of the synthesizing data.


\label{chap: Methodology}
\section{Proposed Method}
\label{sec:proposedmethod}

\begin{figure}[!ht]
    \centering
    \includegraphics[width=10cm]{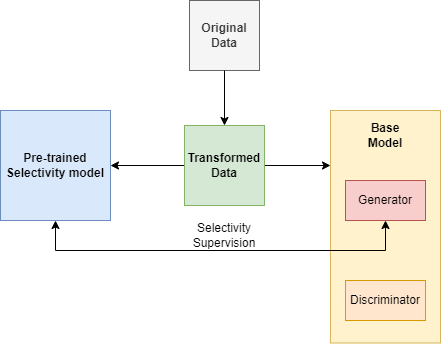}
    \caption{Methodology Overview}
    \label{fig:Methodology}
\end{figure}

We use the proposed methodology to enhance the Base Model to fulfil query selectivity constraints through the Pre-Trained Selectivity Model. The Base Model could be any existing GAN-based tabular data generation model. \fref{fig:Methodology} shows the overview of the method process. The whole process can break into three steps:
\begin{enumerate}
    \item Data Transforming: Make the GAN model compatible with the mixed data type.
    \item Pre-trained Selectivity Model: Provide supervision for generated data.
    \item Base model Training: Incorporates the selective score during the Base Model training. 
\end{enumerate}

\subsection{Data Transforming}
Tabular data usually contains multiple columns with continuous mixed type and categorical type. From \cref{chap: Related Work} we know that generating tabular data with mixed data types could be challenging for some of the existing GAN-based methods.  \texttt{CTGAN} uses a Reversible Data Transforms(RDT) for data pre-processing to handle mixed data types. The RDT converts the categorical features to one-hot vectors and uses the "Mode-Specific Normalization" method to convert a continuous feature to the corresponding Mode and normalized value. 

\subsubsection*{Categorical Data}

There are two types of categorical data in tabular data, ordinal and nominal. Nominal data is classified without a natural order or rank (for example: Male and Female), whereas ordinal data has a predetermined or natural order (for example: Small, Medium and Large). \texttt{CTGAN} uses One-hot Encoding to convert both nominal and ordinal data. \textbf{One-hot Encoding} is a simple encoding method to creates additional features based on the number of unique values in the categorical feature. This encoding method usually used to convert nominal type categorical data, but it may not suitable for ordinal data. One-hot Encoding may not catch the natural ordered relationship between each response for ordinal data.  For example, if we have a nominal categorical variable $D_{
gender}$ and an ordinal categorical variable $D_{size}$. One-hot Encoding represents $D_{gender}= \{F, M\}  \text{ are } [1,0] \text{ and } [0,1]$, represents $D_{size} = \{small, medium, large\} \text{ are } [1,0,0]\text{, }[0,1,0]\text{ and } [0,0,1]$. Ont-hot Encoding provides an equal distance between outcomes. However, for ordinal variable $D_{size}$ the distance between $small$ and $large$ should be larger than the distance between $small$ and $medium$ since the order matters in ordinal variables. The One-hot Encoding loses natural order information when converts ordinal variables.
\\\\As a result, we decide to use \textbf{Ordinal Encoding} for ordinal variables to maintain the inner ordered relations to make sure the GAN model can understand and harness this relationship during the generation step. By using the Ordinal Encoding, the representation of $D_{size} = \{small, medium, large\} \text{ is } [1,2,3]$. We use $d_{i,j}$ to represent the converted value from the ith categorical column and jth row.

\subsubsection*{Continuous Data}
\begin{figure}[htbp]
    \centering
    \includegraphics[width=10cm]{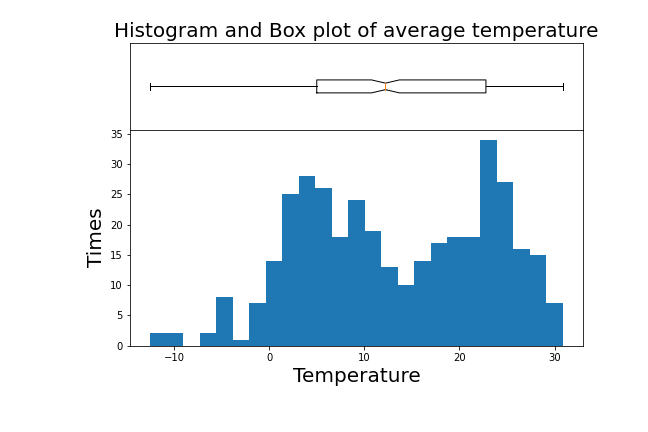}
    \caption{Example of multi-modal distribution}
    \label{fig:multi-modal distribution}
\end{figure}

We continuous data we used the Mode-specific normalization method, which has been introduced in \texttt{CTGAN} from session \ref{sec:ctgan}. The Mode-specific normalization method normalizes a continuous column with multi-modal distributions. \fref{fig:multi-modal distribution} shows the histogram of daily temperature records through one year. It is a typical two-modal distribution example where the two distribution indicates the temperature in winter and summer.

To be specific,the Mode-specific normalization uses variational Gaussian mixture(VGM) model to find the number of Modes for each continuous column $C_i$ and the parameters mean $\eta_k$ and standard deviation $\phi_k$ of each mode. Then, for each value $c_{i,j}$ in $C_i$ compute the probability $\rho_k$ of $c_{i,j}$ coming from each mode. Using the $\rho_k$ to choose which mode the value belong to and using the parameters from the selected mode to normalized the $c_{i,j}$. Finally, record the normalized value $\alpha_{i,j}$ and the chosen mode represent in one-hot vector $\beta_{i,j}$.

\begin{figure}[htp]
    \centering
    \includegraphics[width=12cm]{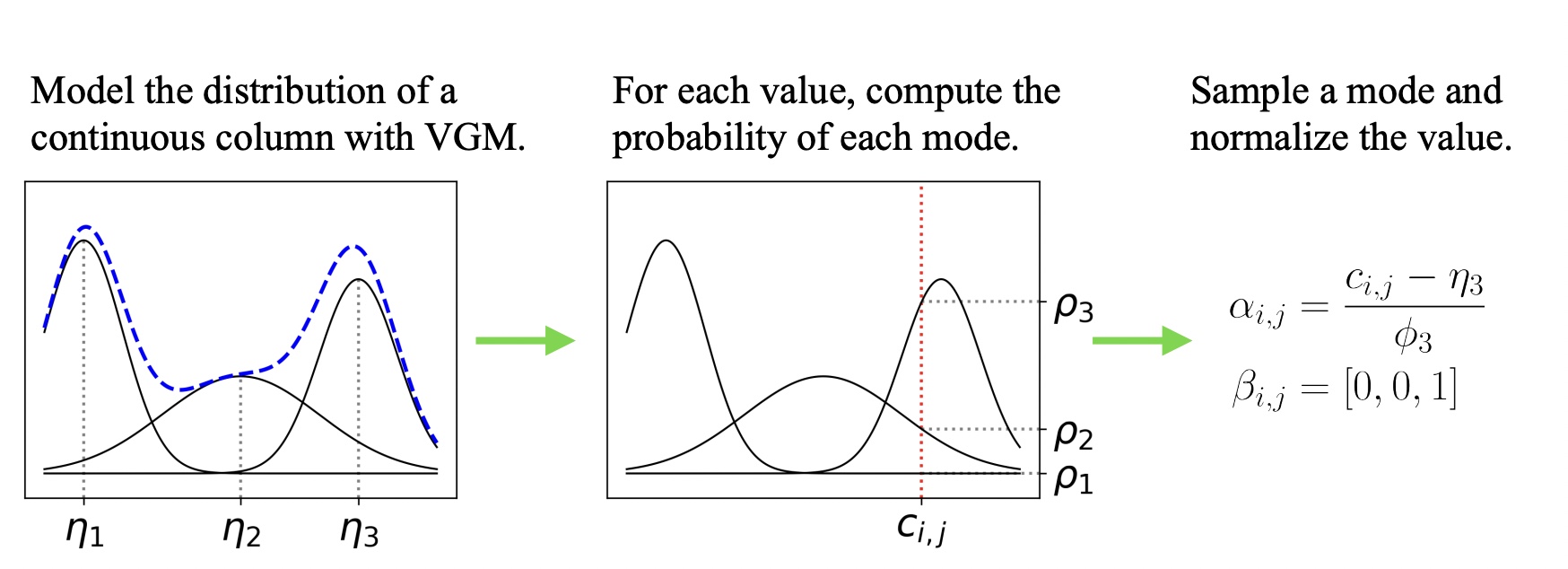}
    \caption{Example of mode-specific normalization}
    \label{fig:Mode}
\end{figure}

For example, in \fref{fig:Mode} there are three mode are found by VGM with the mean value of $\eta_1, \eta_2 \text{ and } \eta_3$. Each mode is a Gaussian distribution with the parameter mean $\eta_k$ and standard deviation $\phi_k$. Then a continuous value $c_{i,j}$ from column $C_i$ appears. $c_{i,j}$ has the $\rho_1$, $\rho_2$ and $\rho_3$ of coming from each mode. It is more likely to come from the third mode $\eta_3$. Thus, $c_{i,j}$ is normalized by the parameters from the third mode, using the formulae: $$\frac{c_{i,j}-\eta_k}{\phi_k}.$$ We use $\alpha_{i,j}$ to represent the normalized value, $\beta_{i,j}$(Mode indicator) is a one-hot vector which represents it allocated into the third mode $[0,0,1]$. Finally, a continuous value $c_{i,j}$ can be converted into the concatenate of $\alpha_{i,j}$ and $\beta_{i,j}$

After the above transformation, the representation of the jth row becomes the concatenation of continuous and discrete columns can be written as follows:
$$
r_j = \alpha_{1,j}\oplus \beta_{i,j} \oplus ...\alpha_{N_c,j}\oplus \beta_{N_c,j} \oplus d_{1,j}...\oplus d_{N_d,j}
$$
A modified Reversible Data Transformer(mRDT) is built to transform the origin data into the processed data and convert the processed back to the original form.

\subsection{Pre-trained Selectivity Model}
In the second step, we need to train a model to estimate the selectivity for the original tabular data. After step one, the dimension of transformed data increases a lot. Therefore, we need a selectivity estimation model that can handle high-dimensional data. Finally, We decided to employ the simple version of \texttt{Selnet}~\cite{selnet} as our Pre-trained Selective Model. \texttt{Selnet} is designed to estimate the selectivity for high-dimensional data, which is quite suitable for our case. It is a regression-based deep learning model that learns a query-dependent piecewise linear function as a selectivity estimator. The original implementation of \texttt{Selnet} contains a Data Partitioning part to improve the accuracy of estimation on large-scale datasets. We drop this part for the current implementation.

Recall the \textbf{Definition \ref{def:selectiviy} Selectivity} from Section \ref{sec:selnet}:\\
Given a database with $d$ dimensional vector $\mathbb{D}=\{\textbf{o}_i\}^{n}_{i = 1}$, $\textbf{o}_i \in \mathbb{R}^d$. Provide a distance function $dist(\,\cdot\,)$, a scaler threshold $t$ and a query object $x\in\mathbb{R}^d$. The estimate the selectivity in the database can be written as: $$|\{\,\textbf{o}\,|\,dist(\textbf{x},o)\le t,\textbf{o}\in \mathbb{D}\}|$$
The selectivity (i.e., the ground truth label) $y$ of a query object $\textbf{x} $ and a threshold $t$ as generated by a \textbf{value function}:
$$y = f(\textbf{x},t,\mathbb{D})$$
Through the \textbf{Definition \ref{def:Threshold} Threshold Partitioning}, the estimation of selectivity can be written as Equation \eqref{eqn:threshold}. We use this Threshold Partitioning method, which means $L+1$ piece-wise linear function is used to implement the interpolant function $g_i(x,t)$. 
Each $g(x,t)$ contains corresponding $\tau_i\text{ and }p_i$. The final estimation function can be re-parameterized as $\hat{f}(\textbf{x},t,\mathbb{D};\Theta)\text{ where }\Theta\text{ represents } \{(\tau_i,p_i)\}^{L+1}_{i=0}$. Therefore, we need to find the best control points$\tau$ and estimate the selectivity $p$, then combine all the $\tau\text{ and }p$ to calculate the total selectivity $\hat{y}$. 
\begin{figure}[htp]
    \centering
    \includegraphics[width=12cm]{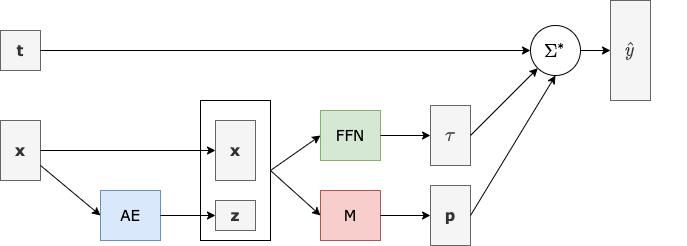}
    \caption{Architecture of Pre-trained Selectivity Model}
    \label{fig:selnet}
\end{figure}
\\\\\fref{fig:selnet} shows the architecture of the Pre-trained Selectivity Model. It is a complicated model which combines three deep neural network components. 
We firstly input the query objects \textbf{x} to an Autoencoder(\textbf{AE}) to learn a latent representation. The \textbf{AE} encourages the model to use latent representation and query distributions for the piecewise linear function. The AE makes the model more generalized and better for handling query objects beyond the training data. Then the query objects \textbf{x} is feed in to the \textbf{AE} to learn the representation \textbf{z}.
The origin \textbf{x} and representation  \textbf{z} is concatenated together to form $[x;z]$. After that the $[x;z]$ is fed into two independent deep neural networks: a feed-forward network (\textbf{FFN}) and \textbf{M}. The \textbf{M} is a encoder-decoder model. In the encoder, an FFN is used to generate (L + 2) embeddings:
$$
[h_0; h_1; . . . ; h_{L+1}] = \text{FFN}( [x; z]),
$$ where $h_i$s are high-dimensional representations to represent the latent information of $\textbf{p}$. Then, we adopt adopt (L + 2) linear transformations with the ReLU activation function:
$$
k_i = \text{ReLu}(w_i^Th_i+b_i)
$$
Then, we have $\textbf{p} = [k_0,k_0+k_1, . . . \sum^{L+1}_{i=0}k_i]$. The output of \textbf{FFN} and \textbf{M} can be converted to the $\tau$ and $\textbf{p}$ vector. Finally they can combine with the threshold $t$ and fed into the operator $\Sigma^*$ to compute the estimated selectivity $\hat{y}$. The Estimation Loss is used to estimate the loss between the true selectivity $y$ and the estimated value
$\hat{y}$ of a query $(x, t)$.
$$
    J_{\text{est}}(\hat{j}) = \sum_{((x,t),y)\in \mathcal{T}_{\text{train}}} l(f(x,t,\mathbb{D}),\hat{f}(x,t,\mathbb{D})) = l(y,\hat{y})
$$
Due to the use of \textbf{AE}, the final loss function\eqref{eqn:selnet} is a linear combination of the Estimation loss and the loss of the \textbf{AE}($J_{AE}$) during the training.
\begin{equation}
    J(\hat{f}) = J_{\text{est}}(\hat{j}) + \lambda \,\cdot\, J_{AE}
    \label{eqn:selnet}
\end{equation}

In our setting, the \texttt{Selnet} is trained to make sure the GAN model satisfies the selectivity constraint. We denote the transformed $T_{origin}$ through mRDT as $\mathbb{D}_{origin}$. Using the entire data set as the training query objects $\textbf{Q}_{\text{train}}$ to ensure the \texttt{Selnet} can be trained properly. Then, the labels $\textbf{y}_{\text{train}}$ and thresholds $\textbf{t}_{\text{train}}$ are generated based on $\mathbb{D}_{origin}$. Then, we use the training query objects $\textbf{Q}_{\text{train}}$ and $\textbf{y}_{\text{train}}$  to train the selectivity model. After the training is done, the model is ready to evaluate the performance of any arbitrary synthesizing data through Mean Squared Error (MSE). The evaluation result can be written as follows:
\begin{equation}
    \mathcal{L}_{Sel} = \text{MSE}(y,\hat{y})
    \label{eqn:sel_loss}
\end{equation}

\subsection{Base Model Training}
\begin{figure}[htp]
    \centering
    \includegraphics[width=13cm]{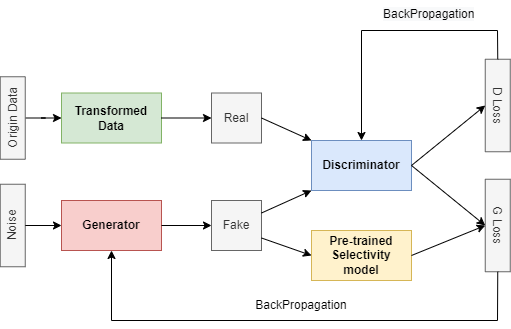}
    \caption{Base Model Training Process}
    \label{fig:trainselGAN}
\end{figure}

After the Selectivity Model is trained, we use the trained selectivity model to enhance the Base Model. The Base Model could be modified from any existing GAN-based tabular data generation model. \fref{fig:trainselGAN} shows the flowchart of the training process of Base Model. 
\\
In a standard GAN model, the Generator $\mathcal{G}$ will generate Fake data each iteration after receiving a random noise. Then the Real data from the original input and Fake data will both send into the Discriminator $\mathcal{D}$. The $\mathcal{D}$ recognizes the Fake and Real, and then the feedback is produced to Generator Loss and Discriminator Loss, respectively. After that, the loss will be back-propagated to the $\mathcal{G}$ and $\mathcal{D}$ and start the next iteration. In our method,
we can pick any arbitrary standard GAN model as the Base Model and then make two changes during the Base Model training so that the produced data from Base Model can satisfy our requirements. One is the Selectivity Evaluation for the Fake data, and the other is the Generator Loss Function Modification before back-propagation.

\subsubsection*{Selectivity Evaluation}
During the GAN training process, the generator synthesis a fake tabular data in each iteration. The synthesised data has the same format and dimension with the transformed data, therefore there is no need to further per-processed the synthesised data. Then for each Fake data, we generate the test query $\textbf{Q}_{\text{test}}$, labels $\textbf{y}_{\text{test}}$ and thresholds $\textbf{t}_{\text{test}}$. Note that the labels $\textbf{y}_{\text{test}}$ are
computed on $\mathbb{D}_{\text{origin}}$, not $\textbf{Q}_{\text{test}}$, thus the estimated selectivity performance will not be overestimated. Then the $\textbf{Q}_{\text{test}}$ and labels $\textbf{y}_{\text{test}}$ will send to the Pre-trained Selectivity Model to predict the $\hat{\textbf{y}}_{\text{test}}$. The evaluation metric is MSE which is calculated through equation \eqref{eqn:sel_loss}. The evaluation result is donated as $\mathcal{L}_{Sel}$. The $\mathcal{L}_{Sel}$ indicates that if the $\textbf{Q}_{\text{test}}$ fulfill the selectivity constraints, the less MSE score means the better performance. 

\subsubsection*{Generator Loss Function Modification}
The selectivity estimation score $\mathcal{L}_{Sel}$ could not only indicate the performance of the Fake data but also shows the ability of the Generator $\mathcal{G}$ in the current status. To improve the capability of $\mathcal{G}$, we add the $\mathcal{L}_{Sel}$ to the Generator Loss $\mathcal{L}_\mathcal{G}$. Thus the $\mathcal{G}$ will modify itself to minimize the loss function then the selectivity constraints will be satisfied. Thus, the loss function for the Generator $\mathcal{G}$ is adapted as:
\begin{equation}
     \mathcal{L}_\mathcal{G}^* = \mathcal{L}_\mathcal{G} + \alpha\cdot \mathcal{L}_{Sel}
     \label{eqn:selgan_Gloss}
\end{equation}
where the $\mathcal{L}_{Sel}$ is the Selectivity Loss \eqref{eqn:sel_loss} and $\alpha$ is the hyper-parameter
indicates the weight of the Selectivity Loss term to avoid the large selectivity Loss dominating the whole loss value. For the Discriminator $\mathcal{D}$ we remain the same loss function. In general, the loss functions should be globally continuous and differentiable. Fortunately, the added term to Generator loss is MSE, one of the simplest and most typical loss functions. Thus the modified loss function should work well theoretically.

Overall, Algorithm \ref{alg:selganTraining} shows the Generator training algorithm.

\begin{algorithm}

\textbf{Input}: $T_{origin}$\\
\textbf{Output}: Trained Generator $\mathcal{G}$
\begin{algorithmic}[1]

\State $\texttt{Selnet} \gets \text{Pre-Trained Selectivity model}$
\State $\mathcal{G} \gets \text{Generator}$
\State $\mathcal{D} \gets \text{Discriminator}$

\State $\mathbb{D}_{origin} \gets \text{mRDT}(T_{origin})$
\For{number of epoch}
    \For{k steps}
        \State {Create a mini-batch of random noise $Z = \{z_1,...,z_n\}$}
        \State Sample a mini-batch of real data $X = \{X_1,...,X_n\}$ from $\mathbb{D}_{origin}$ 
    
        \State Train $\mathcal{D}$ by maximizing equation \eqref{eqn:DLoss}
    \EndFor
    \State Create a mini-batch of random noise $Z = \{z_1,...,z_n\}$
    \State Generate $\textbf{Q}_{\text{test}}$ and Labels $\textbf{y}_{\text{test}}$ using $\mathcal{G}(Z)$ \Comment{Labels are computed on $\mathbb{D}_{\text{origin}}$}
    \State $\mathcal{L}_{Sel} \gets \texttt{Selnet}([\textbf{Q}_{\text{test}}:\textbf{y}_{\text{test}}])$
    \State $\mathcal{L}_\mathcal{G}^* = \mathcal{L}_\mathcal{G} + \alpha\cdot \mathcal{L}_{Sel}$
    \State Train $\mathcal{G}$ by minimizing $\mathcal{L}_\mathcal{G}^*$
\EndFor
\State \Return $\mathcal{G}$

\end{algorithmic}
\caption{Generator Training Algorithm}\label{alg:selganTraining}
\end{algorithm}

\subsection{Base Model}
\label{sce:basemodel}
This thesis employs two existing models as our Base Model to test our proposed method's compatibility. 
\subsubsection*{CTGAN}

\texttt{CTGAN} is a commonly used baseline tabular data synthesizing GAN model. It could synthesize data in high quality and solve imbalanced data problems by introducing additional conditions. It uses RDT to prepossess data and then send the prepossessed data to the \texttt{CTGAN} model.  \texttt{CTGAN} uses PACGAN framework to prevent mode collapse and WGAN loss \eqref{eqn:critic} with gradient penalty and Adam optimizer. \fref{fig:ctgan} shows the architecture of \texttt{CTGAN}. We use CTGAN as our Base Model and combine it with our method. The completed model we name it \texttt{SelGAN}.

\subsubsection*{Daisy}
\texttt{Daisy}~\cite{daisy} is a survey paper which compares the different GAN-based frameworks in the tabular data synthesizing field. We want to use a standard GAN-based tabular data generation method to conduct further ablation studies to test whether our proposed method works well. In \texttt{Daisy} work, we find that utilizing a sequence generation mechanism (such as recurrent neural networks(RNN) and long
short-term memory (LSTM)) as Generator could generate attributes separately in sequential timesteps and provide robust results. As a result, we decide to employ an LSTM as the Generator and a standard MLP as the Discriminator as the Base Model. Again, we use RDT to prepossess data and then send the prepossessed data to the Base Model and add our proposed method. The final resulted model we call it \texttt{Daisy-sel}.


\clearpage   

\lhead{\emph{Chapter 4 Experiments}}

\chapter{Experiments}
In this chapter, we will discuss the implementation of experiments. We will briefly introduce the five real-world datasets in section \ref{sec:Dataset}. Section \ref{sec:baselinemodels} and Section \ref{sec:parametersetting} will talk about the four baseline models and parameter setting for each models and dataset. Section \ref{sec:evaluationresult} shows the evaluation results using the metrics mentioned in section \ref{sec:evaluationmetrics}. An ablation study is also conducted to demonstrate the efficacy of our method.

\label{chap: Experiments}

\section{Dataset}
\label{sec:Dataset}
In our experiments, we choose five commonly used real-world from UCI Machine Learning Repository and Kaggle. 
\begin{itemize}
    \item Adult\tablefootnote{Adult: \url{http://archive.ics.uci.edu/ml/datasets/adult}}: Contains the work-hour attribute has the information of work hours per week for each individual.
    \item Covertype\tablefootnote{Covertype: \url{http://archive.ics.uci.edu/ml/datasets/covertype}}: Contains cartographic variables for four wilderness areas located in the Roosevelt National Forest of northern Colorado. 
    \item Ticket\tablefootnote{Ticket: \url{https://www.transtats.bts.gov/DataIndex.asp.}}: Contains the data for the plane tickets.
    \item News\tablefootnote{News: \url{https://archive.ics.uci.edu/ml/datasets/online+news+popularity}}: Contains a heterogeneous set of features about articles published. 
    \item CreditCard\tablefootnote{CreditCard: \url{https://www.kaggle.com/mlg-ulb/creditcardfraud}}: Contains PCA data with 28 dimensions of fraudulent credit card transactions information. 

\end{itemize}

Those data sets contains both high and low dimension data with mixed type which means all dataset contain both numerical and categorical data. \tref{tab:dataset_summary} shows the summary of chosen datasets. As a result, all data can be sent to both regression and classification task in the later machine learning utility tests. For regression task we use \texttt{educutation\_num} from Adult, \texttt{soil\_type} from Covertype, \texttt{amount} from CreditCard, \texttt{shares} from News,  and \texttt{passengers} from Ticket as our response. For classification task we use \texttt{sex}' from Adult, \texttt{cover\_type} from Covertype, \texttt{class} from CreditCard, \texttt{is\_lifestyle} from News,  and \texttt{mktcoupons} from Ticket as our response.

\begin{table}[h]
    \centering
    \begin{tabular}{c|c|c|c|c|c}
    \hline
        Name & \#Instance & \#Columns & \#Continuous & \#Ordinal & \#Nominal  \\\hline\hline
        Adult & 30148 & 15 & 6 & 1 & 8  \\
        Covertype & 77469 & 13 & 10 & 0 & 3  \\
        Ticket  & 5000 & 38 & 4 & 0 & 34  \\
        News & 39644 & 60 & 46 & 0 & 14  \\
        CreditCard & 14241 & 30 & 28 & 0 & 1 \\
        \hline
    \end{tabular}
    \caption{Summary of datasets}
    \label{tab:dataset_summary}
\end{table}

\section{Baseline Models}
\label{sec:baselinemodels}
For baseline models, we mentioned existing works VAE, \texttt{MedGAN}, \texttt{tablgeGAN}, \texttt{CTGAN}, and \texttt{OCTGAN} in \cref{chap: Related Work}. \texttt{MedGAN} is limited to generated binary response which is not suitable for our data set. Therefore, we abandon the \texttt{MedGAN}. \texttt{OCTGAN} failed in Ticket due to mode collapse.
\texttt{tablgeGAN} has a classifier component inside to maintain the semantic, which needs a binary response column as the label input for the classifier. Only Adult and CreditCard datasets contain a binary response. Therefore \texttt{tablgeGAN} only successfully works for those two datasets. Also, generated data could not produce the label input, and we do not adopt \texttt{tablgeGAN} to the Selectivity Estimation Test. \texttt{SelGAN} is one of our resulting model mentioned in \ref{sce:basemodel}. 

\tref{tab:compatible} shows the compatibility for each model.

\begin{table}[h]
    \centering
    \begin{tabular}{c|c|c|c|c|c}
    \hline
        Model & Adult & Covertype & Ticket & News & CreditCard  \\\hline\hline
        \texttt{SelGan} & \checkmark  & \checkmark & \checkmark & \checkmark & \checkmark  \\
        \texttt{CT-GAN}& \checkmark & \checkmark & \checkmark & \checkmark &   \checkmark\\
        \texttt{OCT-GAN} & \checkmark &\checkmark  &  &\checkmark  &  \checkmark \\
        \texttt{tablGAN} & \checkmark &  &  &  &  \checkmark \\
        \texttt{VAE} & \checkmark & \checkmark & \checkmark & \checkmark &  \checkmark \\\hline
    \end{tabular}
    \caption{Model Compatible table }
    \label{tab:compatible}
\end{table}

\section{Evaluation Metrics}
\label{sec:evaluationmetrics}

Quality assessment of generated data is not a simple task. Through different propose and setting, different evaluation metric is designed. 
We will introduce the designed metric and how to evaluate the synthesizing data in the three aspects.

\subsection{Mode Collapse}

Mode Collapse is a common problem for the GAN-based model. The synthesizing data is supposed to be diverse. Since the generator is always trying to find the one output that seems most plausible to the discriminator, thus if a generator produces a plausible output, it might learn only to produce that output. Therefore generators keep producing a small set of output over and over again. We will input the same amount of origin data for each model by testing this problem and letting them generate the same amount of data. Check if there are duplicated records exist in the generated data.

\subsection{Visualization}
We expect the synthesized data to be close to the original data. For numerical columns, one simplest way is to visualize the data distribution. Visualizations could provide us with a clear view of the comparison results. We could recognize if the model generates the correct number of modes or if the model can handle outliers. We compared the cumulative distribution functions (CDFs) of each column's origin data and synthesized data to evaluate whether a generated synthesized data is close to the origin data statistically. Also, we plot Pearson correlation heat maps. A correlation heat map is a graphical representation of a correlation matrix representing the correlation between different variables. These Pearson Correlation Heat Maps can test if the synthetic data contains the inner linear correlation between features.

\subsection{Selectivity Estimation}
As one of the major contributions of our works, we will evaluate the selectivity estimation score for $T_{origin}$ and $T_{Synth}$. This test is conducted to compare the ability to handle selectivity for models. Similar to before, we used a pre-trained selectivity estimation model metric. Each $T_{Synth}$ will generate the 1000 queries $\textbf{Q}_{\text{Synth}}$ based on $\mathbb{D}_{origin}$. The queries are sent into the pre-trained model and result in MSE score to indicate if the $T_{Synth}$ satisfy the selectivity constraints. Each experiment is repeated ten times. The average lower MSE score shows better performance on fulfilling selectivity constraints.

\subsection{Machine Learning Utility}

Data utility is highly dependent on the specific needs of the synthetic data for down-streaming tasks. We use the Machine learning score to evaluate the effectiveness of using synthetic data as training data for machine learning tasks.
We conduct both supervised learning tasks and unsupervised learning to evaluate the data quality thoroughly. For the supervised learning task, we first need to choose our task and separate the features $X$ and labels $y$ from the original data table $T_{origin}$. Then split the origin data feature $X$ into the training set $X_{train}$ and test set $X_{test}$. Then, we extract the features from synthesizing data table $T_{Synth}$ denote as $X_{Synth}$. We use $X_{train}$ and $X_{Synth}$ to train the machine learning models. Then, we could use $X_{test}$ to test the performance of each model. XGBoost, RandomForest and SVM are employed as regressors and classifiers to make the results more accurate. For the classification task, the accuracy or F1 score will be used. In the regression task, $R^2$ and MSE will be used.  

\section{Parameter Setting}
\label{sec:parametersetting}
All experiments are conducted on the `Spartan'~\cite{Spartan} server, which is the general purpose High-Performance Computing (HPC) system operated by Research Computing Services at The University of Melbourne. The partition \textbf{deeplearn} we used contains 13 nodes, each with four NVIDIA V100 graphics cards and six nodes each with four NVIDIA A100 graphics cards. The implementation for the Python version is 3.7.4 (compatible with TensorFlow 1.15, which is required for \texttt{tableGAN}) and 3.8.6. 

We use source code for baseline models \texttt{CT-GAN}\footnote{CT-GAN: \url{https://github.com/sdv-dev/CTGAN}}, \texttt{OCT-GAN}\footnote{OCT-GAN: \url{https://github.com/bigdyl-yonsei/OCTGAN}}, \texttt{tableGAN}\footnote{table-GAN \url{https://github.com/mahmoodm2/tableGAN}}, \texttt{VAE} and \texttt{Daisy}\footnote{VAE and Daisy \url{https://github.com/ruclty/Daisy}} with their default parameters.  

As our method is orthogonal to any existing GAN model, the performance of our method is largely dependent on the capability of the Base Model, and we do not need to do the parameter truing for the Base Model. \texttt{SelGAN} and \texttt{Daisy-Sel} uses the same parameters with \texttt{CTGAN} and \texttt{Daisy}. To ensure the fairness of experiments, we use batch size 500 for VAE and all GAN-based models with 300 training epochs using an Adam optimizer. 

\texttt{Selnet}\footnote{Selnet:\url{https://github.com/yyssl88/SelNet-Estimation}} is used as the pre-trained selectivity component. We use complete origin data $T_{origin}$ to train \texttt{Selnet} models with batch size 512. The training epoch for \textbf{AE} is 100 and 120 for other models. The parameter $\alpha$ from equation \eqref{eqn:selgan_Gloss} is set as 0.01 to avoid large selectivity loss dominate the whole loss value.

\section{Results analysis}
In this session, we conduct a complete evaluation of our method to understand the quality of synthetic data in the metrics mentioned before.
\label{sec:evaluationresult}

\subsection{Mode Collapse}
\tref{tab:modecollpase} show the rate of repeated data. We use this method to check if the model suffers from mode collapse. The higher value indicates that the mode collapse phenomenon is serious. From the table, we can see that the \texttt{VAE} model suffers from the mode collapse problem the most. \texttt{SelGAN} is not suffering from this problem as well as its Base Model \texttt{CTGAN}. The standard capability of \texttt{SelGAN} depends on the Base Model. We can only comment that our method will not cause any mode collapse, but we could not make any comment on if our method could ease mode collapse.
\begin{table}[htbp]
    \centering
    \begin{tabular}{c|c|c|c|c|c}
    \hline
        Model     & Adult     & Covertype & Ticket & News & CreditCard  \\\hline\hline
        CT-GAN    & 0      & 0  & 0 & 0  & 0   \\
        OCT-GAN   & 0      & 0  & 1   & 0  & 0 \\
        tableGAN  & 3.2    & $-$ & $-$   &  $-$    &   0         \\
        VAE       & 0.779  & 0  & 9.78 & 0  & 84.83  \\\
        SelGAN    &  0     & 0      & 0     & 0       & 0  \\\hline
      
    \end{tabular}
    \caption{Rate of Repeated Data (\%)}
    \label{tab:modecollpase}
\end{table}

\subsection{Visualization}
 Due to the space limit, we do not reveal the complete visualization plots in this chapter, and more plots are shown in \aref{Appendix_figure}.
\\
\subsubsection*{CDF Comparison}
\begin{figure}[htbp]
    \centering
    \includegraphics[width=15cm]{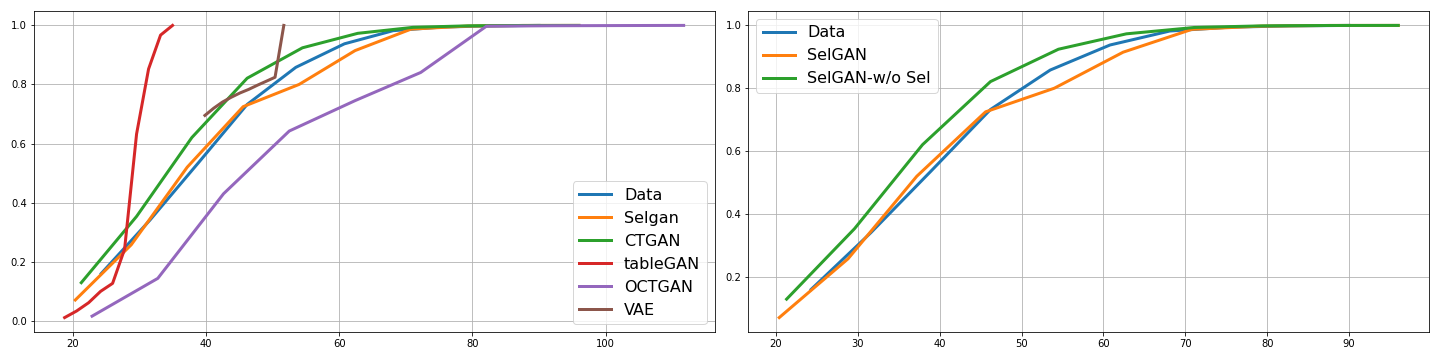}
    \caption{age in Adult}
    \label{fig:age_A}
\end{figure}

\begin{figure}[htbp]
    \centering
    \includegraphics[width=15cm]{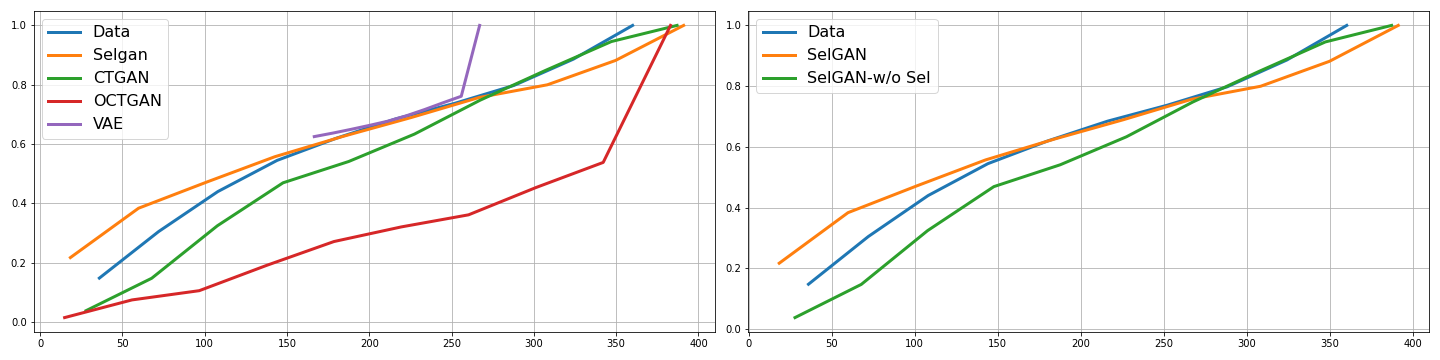}
    \caption{aspect in Covertype}
    \label{fig:aspect_Co}
\end{figure}

\begin{figure}[htbp]
    \centering
    \includegraphics[width=15cm]{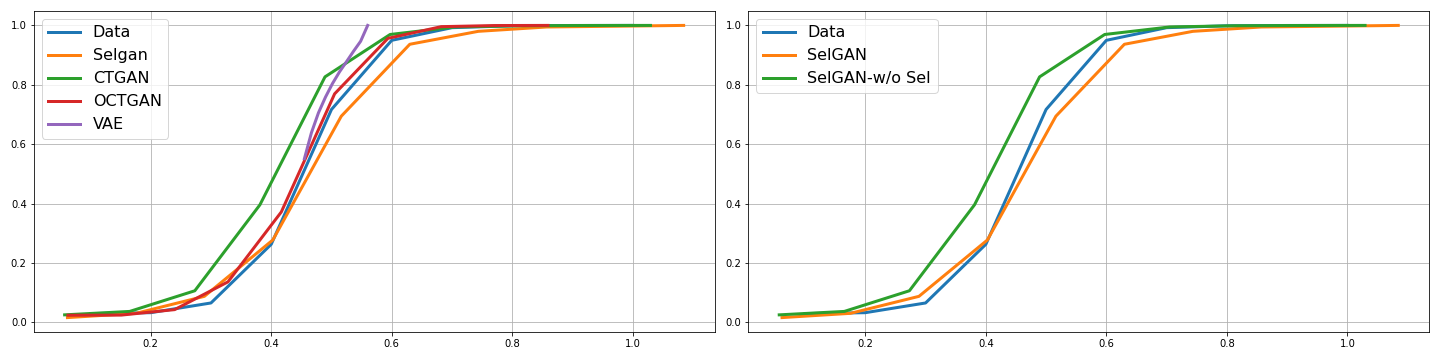}
    \caption{global subjectivity in News}
    \label{fig:global_subjectivity_N}
\end{figure}

 Three interpret-able continuous columns are chosen for the CDF comparison. \fref{fig:age_A}, \fref{fig:aspect_Co} and \fref{fig:global_subjectivity_N} shows \texttt{age} in Adult, \texttt{aspect} in Covertype and \texttt{global\_subjectivity} in news.
 The left-hand side shows an overall CDFs comparison among all baseline model and our resulting model \texttt{SelGAN}. The comparison of CDFs for \texttt{SelGAN} and \texttt{SelGAN-w/o Sel} are showns in the right hand side. The \texttt{SelGAN-w/o Sel} model removes the selectivity estimation component for ablation studies. From thses plots, CDFs of \texttt{SelGAN} in orange are close to CDFs of $T_{origin}$ in blue, suggesting that \texttt{SelGAN} performs quite well. However, \texttt{SelGAN} could not fit well at the beginning and end in \fref{fig:aspect_Co}, that means our \texttt{SelGAN} does not always successfully capture the statistics of the $T_{origin}$. But, \texttt{SelGAN} still outperform among other base line models visually.
 

\subsubsection*{Correlation Heat Map}
\begin{figure}[htbp]
    \centering
    \includegraphics[width=\textwidth]{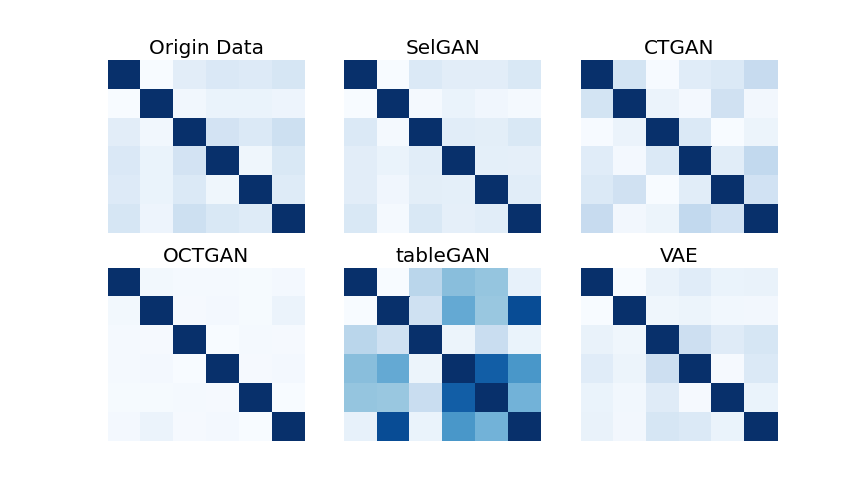}
    \caption{Correlation Heap Map for Adult}
    \label{fig:Corr_A}
\end{figure}

\begin{figure}[htbp]
    \centering
    \includegraphics[width=\textwidth]{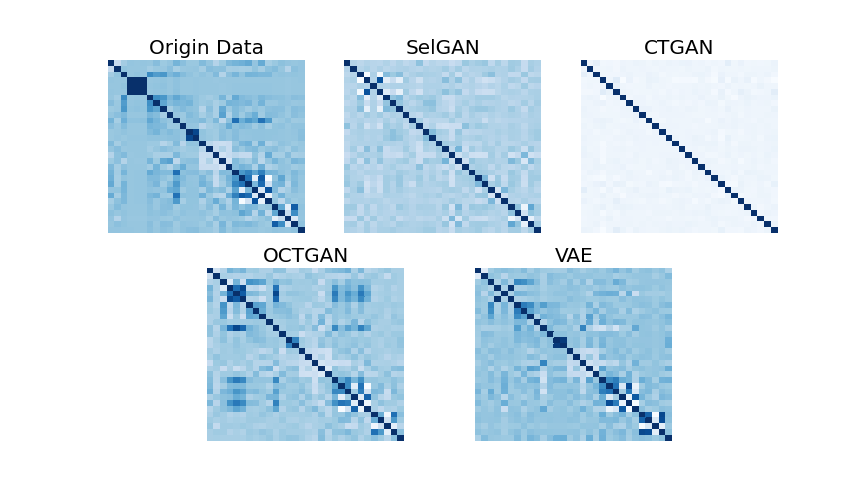}
    \caption{Correlation Heap Map for News}
    \label{fig:Corr_n}
\end{figure}

 \fref{fig:Corr_A} and \fref{fig:Corr_n} shows the Correlation Heap Map for Adult and CreditCard. From \fref{fig:Corr_A}, we can see \texttt{SelGAN} generates a similar pattern and color with the $T_{origin}$. \texttt{CTGAN} and \texttt{VAE} also generate similar patterns and colours, but there are some massive patterns inside. \texttt{OCTGAN} and \texttt{tableGAN} fail to produce the inner correlations. From \fref{fig:Corr_n}, we find \texttt{OCTGAN} and \texttt{VAE} simulate the correlation pretty good, but \texttt{CTGAN} failed. \texttt{SelGAN} produce a similar colour but fails to generate the pattern. We realise that it is hard to evaluate the performance just by visuals. Thus, we conduct a difference in pair-wise correlation comparison test to assess the correlation performance at the quantity level. \tref{tab:map distance} summarize the difference in pair-wise correlation between the correlation matrix of origin data $T_{origin}$ and correlation matrix of synthetic data $T_{Synth}$. The smaller value indicates that the synthetic data can mimic the correlation well. From this table, we find that \texttt{SelGAN} outperforms the other GAN-based models in two datasets. We also investigate an interesting phenomenon, \texttt{VAE} models have the lowest distance in Adult and Covertype, but it does not show ideal results in the other three datasets. However, \texttt{SelGAN} has the second-best results in Adult and Covertype. Thus we can conclude that overall, \texttt{SelGAN} can handle the inner correlation between dataset features well. 

\begin{table}[htbp]
    \centering
    \begin{tabular}{c|c|c|c|c|c}
    \hline
        Model     & Adult     & Covertype & Ticket & News & CreditCard  \\\hline\hline
        
        CT-GAN    & 2.79 & 21.94 & 59.59 & 79.00 & 100.48 \\
        OCT-GAN   & 2.04 & 26.96 &  $-$  & \textbf{46.24} & 85.82  \\
        tableGAN  & 10.09& $-$   &  $-$  & $-$   & 467.84 \\
        VAE       & \textbf{0.76} & \textbf{15.99} & 91.41 & 54.9  & 554.05 \\ \hline \hline
        SelGAN    & 0.93 & 19.79 & \textbf{45.54} & 75.07&  \textbf{66.13}  \\ \hline
      
    \end{tabular}
    \caption{Difference in pair-wise correlation}
    \label{tab:map distance}
\end{table}

\subsection{Selectivity Estimation}
 \tref{tab:selectivityestimation} shows the average summary results for the Selectivity estimation accuracy. From the table, we could see that there are clear reductions for \texttt{SelGAN} among all baseline models. The average decrease rate in MSE score is around $20\%$. Since the ability of our model is also dependent on the ability of the Base Model, we find that the scores are various between baseline models. Thus, further ablation studies are still required to understand better if our method can successfully enhance the existing model to fulfil the selectivity constraints.

\begin{table}[htbp]
    \centering
    \begin{tabular}{c|c|c|c|c|c}
    \hline
        Model     & Adult     & Covertype & Ticket & News & CreditCard  \\\hline\hline
        CT-GAN    & 40.93 & 98.82  & 66.12 & 111.38  & 230.75   \\
        OCT-GAN   & 55.53   & 128.04 &     $-$& 163.73  & 247.22 \\
        VAE       & 63.49   & 97.27  & 68.07 & 132.29  & 238.40  \\
        SelGAN    & \textbf{23.86} & \textbf{82.50}  & \textbf{54.83} & \textbf{104.36}  & \textbf{200.01}   \\\hline
    \end{tabular}
    \caption{Selectivity Estimation MSE in $10^2$}
    \label{tab:selectivityestimation}
\end{table}

\begin{figure}[htbp]
    \centering
    \includegraphics[width=15.5cm]{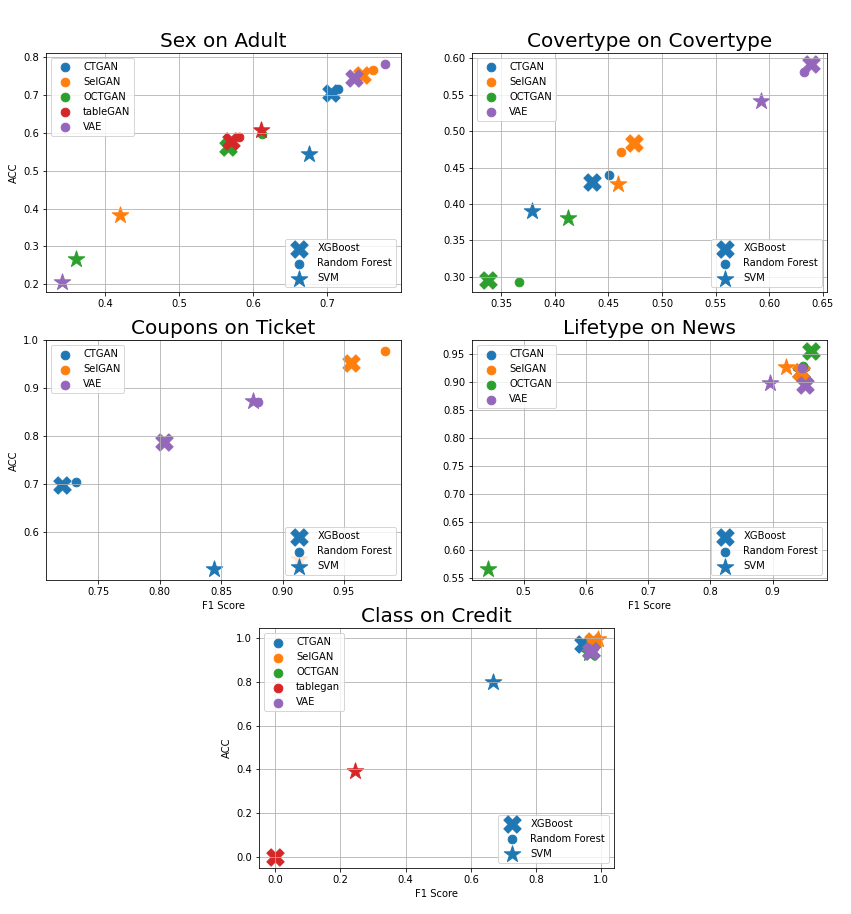}
    \caption{ACC and F1 Score for Classification Task over five datasets}
    \label{fig:accf1}
\end{figure}

\subsection{Machine Learning Utility}

We conducted both classification and regression for all datasets. The predicted labels were mentioned in section \ref{sec:Dataset}. For classification tasks, We plot the Accuracy vs F1-score for all three machine learning models for five datasets (\fref{fig:accf1}). This plot shows that for the dataset Ticket and CreditCard, \texttt{SelGAN} largely outperforms all others across all used machine learning models. For dataset Adult and Covertype, \texttt{SelGAN} outperforms all GAN-based models. For datasets News, the best result and worst both come from \texttt{OCTGAN}, which means the performance of \texttt{OCTGAN} is quite volatile. \texttt{SelGAN} shows relatively stable results with high compatibility.\\ We also use \tref{tab:clf_summary} and \tref{tab:reg_summary} table to summarize the averaged machine learning utility score between $T_{origin}$ and synthetic data in terms of accuracy, F1 score and MSE. A better synthetic data is expected to have a similar performance to the $T_{origin}$. The best evaluation scores have been labelled in boldface. From the two tables, we can see, that mostly
\texttt{SelGAN} outperforms all other GAN-based state-of-the-art
methods in terms of F1-score and MSE. The averaged F1 across the three machine learning models increases up to 6\%, and the averaged MSE decreases up to 20\%.

\begin{table}[htbp]
    \centering
    \begin{tabular}{c|c|c|c|c|c}
    \hline
        Model     & Adult     & Covertype & Ticket & News & CreditCard  \\\hline\hline
        Origin    & 0.66 & 0.69 & 0.98 & 0.96  & 0.99  \\\hline
        CT-GAN    & \textbf{0.65} & 0.42 & 0.80 & 0.91  & 0.91   \\
        OCT-GAN   & 0.47 & 0.32 &  $-$  & 0.81  & 0.93 \\
        tableGAN  & 0.59 & $-$  &  $-$  & $-$  & 0.75           \\
        VAE       & 0.57 & \textbf{0.51} &  0.84 & 0.90  & 0.93  \\\hline
        SelGAN    & 0.63 & 0.42  & \textbf{0.85} & \textbf{0.92}  & \textbf{0.95}   \\\hline
    \end{tabular}
    \caption{Classification Accuracy (F1)}
    \label{tab:clf_summary}
\end{table}
\begin{table}[htbp]]
    \centering
    \begin{tabular}{c|c|c|c|c|c}
    \hline
        Model     & Adult  & Covertype & Ticket & News & CreditCard  \\\hline\hline
        Origin    & 4.17   & 49.03   & 51.67 & 12.03 &4562  \\\hline
        CT-GAN    & 7.66  & 247.05  & 53.62 & 12.62  & 49226     \\
        OCT-GAN   & 9.38   & 116.50  & $-$ & 74.23  & 68724    \\
        tableGAN  & 23.98   & $-$     & $-$    & $-$   & 58099\\
        VAE       & 16.43   & 159.35  & 53.69  & 12.77  & 52410  \\\hline
        SelGAN    & \textbf{5.18} & \textbf{103.27}  & \textbf{53.26} & \textbf{12.28}  & \textbf{39484}   \\\hline

    \end{tabular}
    \caption{Regression Accuracy (MSE) }
    \label{tab:reg_summary}
\end{table}

\section{Ablation Study}
To illustrate the efficiency of our method, we implement an ablation study. Due to the limitation of data resources, from \tref{tab:dataset_summary} we realized there are not many ordinal columns in the experimental datasets. In our proposed method, we use an m-RDT preprocessing method for the proposed data to send to the downstream model more easily. Through m-RDT, we should use Ordinal encoding for ordinal attributes rather than one-hot vectors. That means we will not have enough experimental data to support the impact of using Ordinal encoding or One-Hot encoding. Therefore, we abandon the ablation study of the m-RDT component.\\
As mentioned in Section \ref{sce:basemodel}. We combined our method with two existing GAN base models. These two resulting models are used to test our method's flexibility and compatibility and check if our method can successfully enhance the synthetic data quality. We remove the pre-trained selectivity component from the model and then redo the test to see the performance differences.\\
The following are the two pairs of models we used for the ablation study:
\begin{itemize}
    \item SelGAN and SelGAN-w/o Sel
    \item Daisy-Sel and Daisy
\end{itemize}
On the right-handed side of each figure from \fref{fig:age_A}, \fref{fig:aspect_Co} and \fref{fig:global_subjectivity_N}, the CDS distributions of between $T_{origin}$ and $T_{Synth}$ are revealed. \tref{tab:as-selectivityestimation}, \tref{tab:as-reg} and \tref{tab:as-clf} shows the quantity comparison of selectivity estimation and machine learning performance. In \tref{tab:as-selectivityestimation}, we find that the average decrease rate for adding the selectivity component is 27\%, which is a significant drop in the selectivity score.  That is strong evidence to say that our method does help the current GAN model meet the selectivity constraints. As well as the machine learning utility, we can observe trivial differences among them in most cases. In \tref{tab:as-clf}, \texttt{Daisy-sel} pair in CreditCard dataset and \texttt{SelGAN} pair in Covertype dataset share the same results. 
Therefore, both \texttt{Daisy-sel} and \texttt{SelGAN} show better results than the \texttt{Daisy} and \texttt{SelGAN-w/o Sel} respectively. Considering the high data utility in several datasets, it is crucial to use the selectivity component.
\begin{table}[htbp]
    \centering
    \begin{tabular}{c|c|c|c|c|c}
    \hline
        Model     & Adult     & Covertype & Ticket & News & CreditCard  \\\hline\hline
        Daisy     & 39.77   & 73.86  & 88.83  & 139.99    & 225.22  \\
        Daisy-sel & \textbf{11.21}  & \textbf{67.47}    & \textbf{48.77} & \textbf{82.35}   & \textbf{184.46}  \\\hline \hline
        SelGAN-w/o Sel    & 40.93 & 98.82  & 66.12 & 111.38  & 230.75   \\
        SelGAN    & \textbf{23.86} & \textbf{82.50}  & \textbf{54.83} & \textbf{104.36}  & \textbf{200.01}   \\\hline
    \end{tabular}
    \caption{Ablation Selectivity Estimation MSE in $10^2$}
    \label{tab:as-selectivityestimation}
\end{table}
\begin{table}[htbp]
    \centering
    \begin{tabular}{c|c|c|c|c|c}
    \hline
        Model     & Adult     & Covertype & Ticket & News & CreditCard  \\\hline\hline
        Daisy     & 8.89   & 261.19  & 152.66  & 15.20    & 55135  \\
        Daisy-sel & \textbf{7.83}  & \textbf{143.36}    & \textbf{104.87} & \textbf{12.07}  & \textbf{50798}  \\\hline \hline
        SelGAN-w/o Sel    & 7.66  & 247.05  & 53.62 & 12.62  & 49226     \\
        SelGAN    & \textbf{5.18} & \textbf{103.27}  & \textbf{53.26} & \textbf{12.28}  & \textbf{39484}   \\\hline
    \end{tabular}
    \caption{Ablation Study: Regression Accuracy (MSE) }
    \label{tab:as-reg}
\end{table}
\begin{table}[htbp]
    \centering
    \begin{tabular}{c|c|c|c|c|c}
    \hline
        Model     & Adult     & Covertype & Ticket & News & CreditCard  \\\hline\hline
        Daisy     & 0.54   & 0.47  & 0.77  & 0.72    & 0.97 \\
        Daisy-sel & \textbf{0.63}  &  \textbf{0.52}   & \textbf{0.83} & \textbf{0.75}   & \textbf{0.97}  \\\hline \hline
        SelGAN-w/o Sel    & \textbf{0.65} & 0.42 & 0.80 & 0.91  & 0.91   \\
        SelGAN    & 0.63 & \textbf{0.42}  & \textbf{0.85} & \textbf{0.92}  & \textbf{0.95}   \\\hline
    \end{tabular}
    \caption{Ablation Study: Classification Accuracy (F1)}
    \label{tab:as-clf}
\end{table}

\clearpage   

\lhead{\emph{Chapter 5 Conclusion and Future Work}}

\chapter{Conclusions and Future Work}

\label{chap: Conclusion and Future Work}
\section{Conclusions}

This thesis proposed a flexible method that could enhance any existing  GAN-based tabular data generation model to fulfil the selectivity constraints. Through this method, we can synthesize data with a similar selectivity to the origin data $T_{origin}$. Then the synthetic data can be used to estimate query execution cost more accurately and further estimate the computational resources required if we create queries to the $T_{origin}$. 

In \cref{chap: Introduction}, we introduce the background and aim of our project. We listed four challenges for current tabular data synthesizers, including Data Shortage Issue, Data Privacy Issue, and Data Quality Issues in both the Machine Learning utility and Data Constraints aspects. Then we talked about how the promising tabular data generation method GAN handles these four challenges. We found that current state-of-the-art GAN models have made successes in the first three challenges, but there is a research gap between the current method and the fourth challenge. Motivated by the E-commerce platforms problem, we decided to develop a tabular data generation GAN to model selectivity constraints in tabular data synthesizing and contribute to solving the last Issue.

In \cref{chap: Related Work}, we reviewed some common approaches of tabular data synthesizing methods from statistical and machine learning aspects. Bayesian network~\cite{BN} is a statistical data generation method. It can describe a dataset as a directed acyclic graph. The dependency or relations can be represented through edges and nodes. In the machine learning aspect, we introduce the Variational Auto-encoder. It is a variant of a standard auto-encoder whose encoding distribution is regularised during the training process to keep the approximate features and generate new data. Then, we introduced some GAN variants for tabular data generation including \texttt{MedGAN}, \texttt{tableGAN}, \texttt{CTGAN} and \texttt{OCTGAN}. They developed different methods to make the GAN model solve the first three challenges mentioned before. \texttt{MedGAN} is designed for high-dimensional medical record data, but it can only generate numerical and binary data. \texttt{tableGAN} uses a convolutional neural network as the generator $\mathcal{G}$ and adds a classifier to maintain the semantics of synthetic data. \texttt{CTGAN} and \texttt{OCTGAN} are all aim to solve the in-balanced data distribution problem through a conditional vector and neural ordinary differential equations (NODEs). We also studied different approaches for selectivity estimation, such as the histogram-based and regression-based methods.

In \cref{chap: Methodology}, we talked about the proposed method after consulting many relevant materials and learning from GAN-based tabular data generation methods. Finally, we decided to develop a flexible method to combine any existing GAN-based tabular data generation method to model the selectivity constraints. We first need to send the origin data $T_{origin}$ to train the selectivity estimation model. The pre-trained selectivity estimation model will supervise the Generator's performance during the GAN model training process. The supervision feedback term is added to the Generator loss function, and the Generator can use this feedback to adjust itself in each iteration. We also modified the current data pre-processing step to make the method more suitable for the various data types.

In \cref{chap: Experiments}, we introduce the detail of the experiments. Overall, our proposed method is an enhanced method to the current GAN based model; thus, the major capability in large depends on the Base Model we used to combine. To satisfy the selectivity constraints as well as the quality of the synthetic data, we implement our method to the-state-of-art tabular data generation model \texttt{CTGAN} and resulting \texttt{SelGAN}. The selectivity estimation results show that \texttt{SelGAN} can model tabular data with selectivity constraints successfully, and it is also robust over different datasets to compare with three GAN-based models and one VAE model. Using the results from machine learning utility tests and statistical tests in visual, we can conclude \texttt{SelGAN} can also effectively model tabular data and generate high-quality synthetic data. 
We also combine a GAN model without outstanding performance, resulting \texttt{Daisy-Sel} to test the compatibility of our method. Finally, we conduct an ablation study for \texttt{SelGAN} and \texttt{Daisy-Sel} to analyze the importance of the pre-trained selectivity estimation term. We remove the pre-trained term and rerun the tests. The results show that our method is efficient.

\clearpage

\section{Future Direction}
There are still some limitations of our method. Therefore, in the future work, we plan to explore the following aspects:
\begin{enumerate}

    \item \textbf{More query operators could be considered} \\ There are many query operators in a query execution plan. Now, we only focus on the selection. More commonly used query operators like projection and joint could be considered in the future. 
    
    The selection and projection involve only one single data table. Therefore, the modelling cost for selection and projection should be similar. We use projection queries to train the current model or to replace the current model with a projection estimation model. 
    
    However, modelling joint constraints should be more complicated since the joint operation involves two or more data tables. To model the joint constraints, we consider that we should use multiple parallel GAN models for each data table. Each GAN model will produce a single data table in each iteration, and then we should calculate the joint cost for the current generated table and provide them feedback. Then the feedback is sent to each GAN model to update the $\mathcal{G}$. The multiple GAN training method should be parallel to ensure all sub-GAN models can grow and interact together. One limitation for that the multiple GAN model could be too complicated. Since the number of GAN components depends on the joint constraints, we have to run the same number of sub-GAN components simultaneously if a joint constraint involves a large number of the data table. That could be a huge cost of computational resources.

    \item \textbf{Improvement on Selectivity component} \\ Currently, we use \texttt{Selnet} as our pre-trained selectivity estimation model. We picked this \texttt{Selnet} because it can handle the high-dimensional data. We failed to use the sampling and histograms based or Gradient boosting trees regression-based estimation method due to the curse of dimensionality. In the future, more selectivity estimation models are worth trying to improve estimation accuracy further. To solve the high-dimensional data, we could try to train an Auto-Encoder to learn the representation of synthetic data. We sent the synthetic data into the Encoder to get the representation. Then we sent the representation into the estimation model to receive feedback and use the feedback to update $\mathcal{G}$. Even though the model could not handle the high-dimensional data, it can still handle the representation with lower dimensions. Nevertheless, we need to consider that could the representation have any statistically meaning to fit the logic of estimation models.
    
    In addition, we employed an existing selectivity estimation model previously. Nevertheless, we should have a novel, own-designed selectivity model for our scenario.

    \item \textbf{Satisfy industrial needs} \\
   This project is motivated by the E-commence platform cases. However, our current method is not yet satisfied industrial needs. We have to solve the previous two future directions and then consider them for industry application. There is one big challenge for the E-commence platform in a real-world scenario: we have to dynamically update the users' data. In the E-commence platform cases, users' data or transactions should be updated frequently. Our method can only handle the static data table. If the data table changes, we must redo the GAN model training. However, that should be a common issue for GAN-based models or even all static machine learning models. To solve this problem, we should consider a Dynamic Training method.
    
    \item \textbf{More experiments in other GAN variants}\\ In this thesis, we only test two existing models. These two existing models have different architectures but still can not be representative enough. There are lots of varieties in GAN-based tabular generation methods. We can implement our method to more GAN models and further analyze whether the selectivity improvement would be changed due to the other GAN architectures. 
    
    \item \textbf{Hyper-parameter setting}\\ During the GAN training process, we add the $\mathcal{L}_{Sel}$ into the $\mathcal{L}_{\mathcal{G}}$. We define the parameter $\alpha = 0.01$ to control the weight of $\mathcal{L}_{Sel}$. This weight is to ensure the $\mathcal{L}_{Sel}$ will not dominate the whole loss function. However, the scale of $\mathcal{L}_{Sel}$ for the different datasets is various, and a single value of $\alpha$ may not work well. Thus, more precision experiments are required. We may need to test new $\alpha$ for different datasets or use some normalization function to normalize the  $\mathcal{L}_{Sel}$ in a certain range.

\end{enumerate} 
\clearpage   




\addtocontents{toc}{\vspace{2em}} 

\appendix 
\lhead{\emph{Appendix}}
\chapter{Notations}

\label{Appendix_notation}

\begin{table}[!h]
    \centering
    \begin{tabular}{l l}
    \hline\hline
        Notation & Description  \\ \hline\hline
        General:\\
         $T_{origin}$& Original Data\\
         $T_{Synth}$& Synthetic data produced by model\\\hline
        Pre-processing: \\
         $C_{i}$& ith continuous column\\
         $D_{i}$& ith discrete column\\
        $c_{i,j}$& Value in the i-th continuous column of j-th row\\
         $d_{i,j}$& Value in the i-th discrete column of j-th row \\
        m-RDT& Modified Reversible Data Transforms \\ 
        $\alpha_{i,j}$& normalized value for $c_{i,j}$\\
        $\beta_{i,j}$& Mode representation for $c_{i,j}$ \\
        m-RDT& Modified Reversible Data Transforms \\
        
        \hline
        GAN model: \\
        $\mathcal{D}$& Discriminator \\
         $\mathcal{G}$& Generator\\\hline
         Query Generation: \\\

         $\mathbb{D}_{origin}$& Transformed Original Data\\
        $\mathbf{Q}_{train}$& testing query objects\\         $\mathbf{Q}_{test}$& training query objects\\\hline
         GAN training: \\
         $\mathcal{L}_{\mathcal{G}}$& Generator Loss \\
        $\mathcal{L}_{Sel}$& Selectivity Loss \\
         $\mathcal{L}_{\mathcal{D}}$& Discriminator Loss \\
         $X\sim\{X_1, X_2, X_2\}$& Real data sampled from $\mathbb{D}_{origin}$ \\
         $Z\sim\{Z_1, Z_2, Z_2\}$& Noisy data sampled from $N\sim(0,1)$ \\\hline

    \end{tabular}
    \caption{Notations}
    \label{tab:notations}
\end{table}	

\chapter{Figures}

\label{Appendix_figure}

 Complete CDFs and correlation heat maps visualization plots from \cref{chap: Experiments}

\begin{figure}
    \centering
    \includegraphics[width=\textwidth]{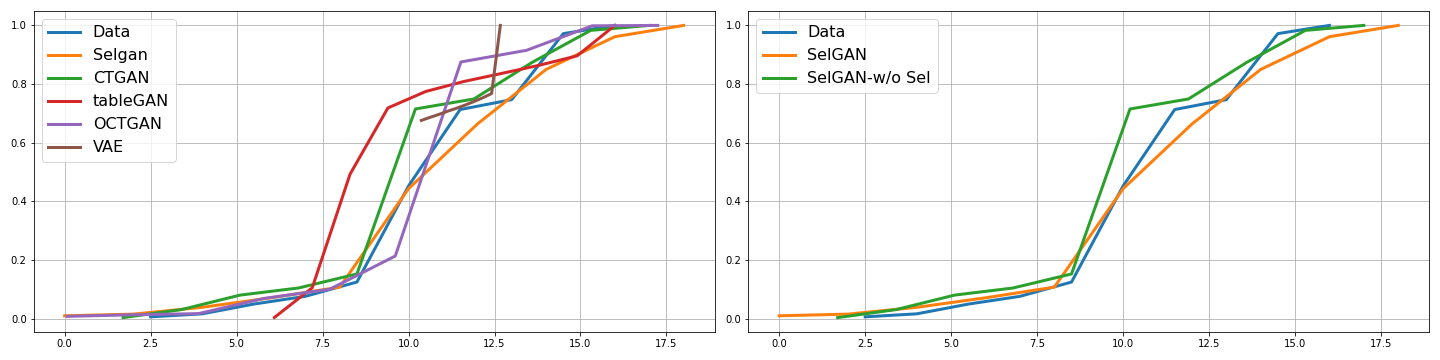}
    \caption{education-num in Adult}
    \label{fig:education-num_A}
\end{figure}

\begin{figure}
    \centering
    \includegraphics[width=\textwidth]{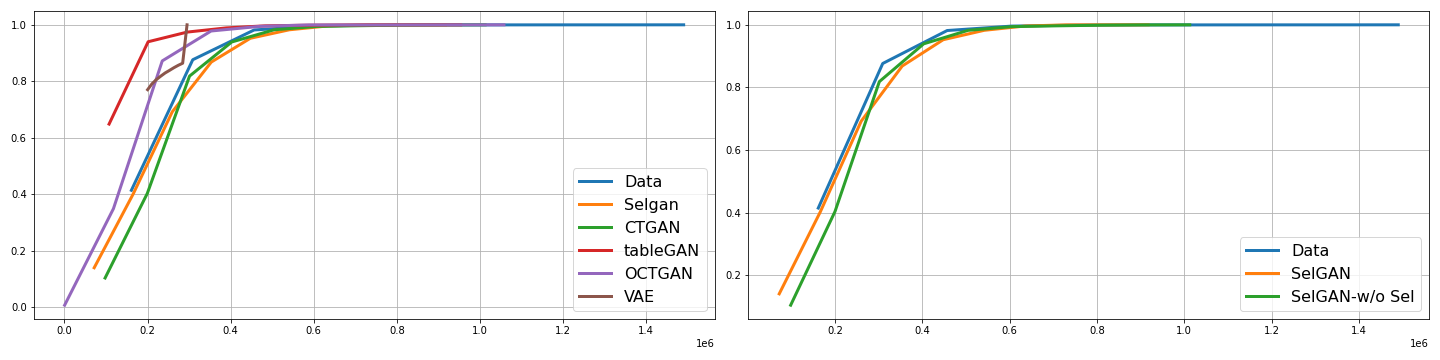}
    \caption{fnlwgt in Adult}
    \label{fig:fnlwgt_A}
\end{figure}

\begin{figure}
    \centering
    \includegraphics[width=\textwidth]{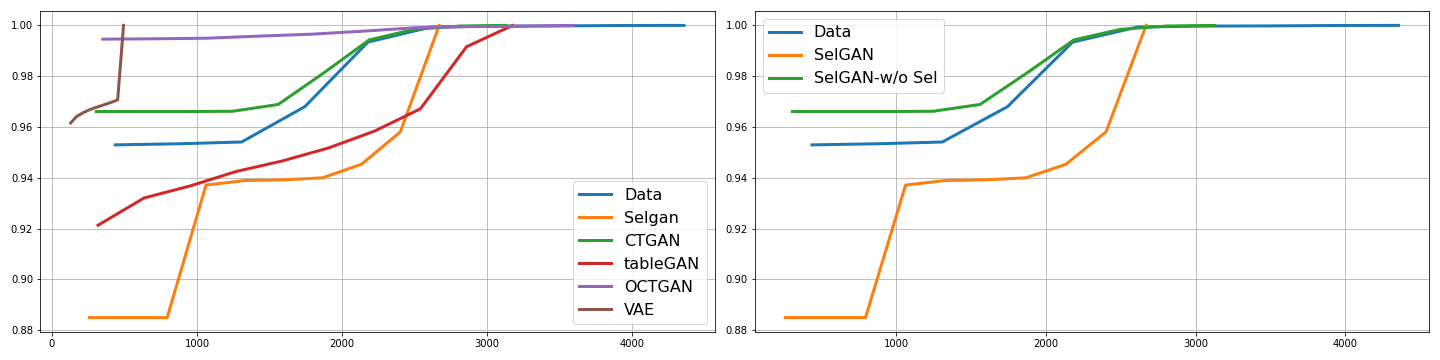}
    \caption{capital-loss in \textbf{Adult}}
    \label{fig:capital-loss_A}
\end{figure}

\begin{figure}
    \centering
    \includegraphics[width=\textwidth]{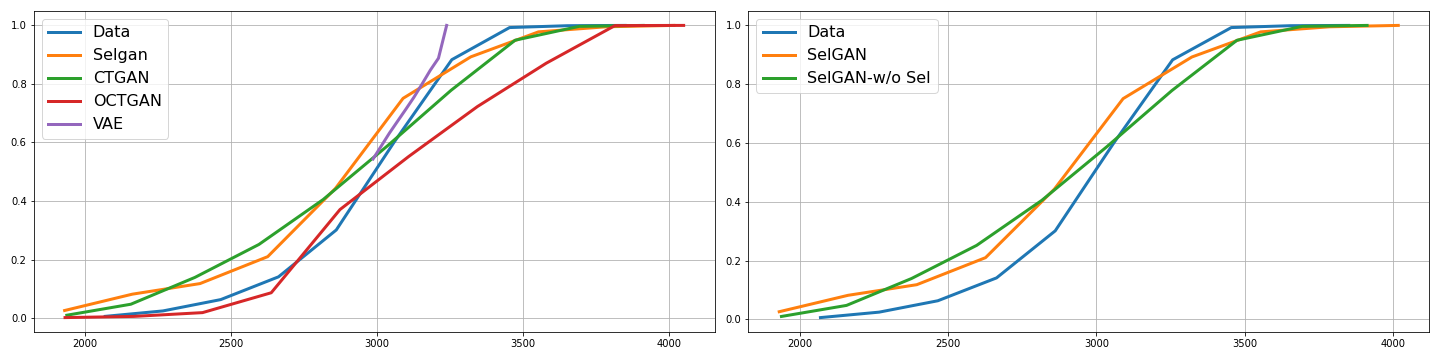}
    \caption{elevation in Covertype}
    \label{fig:elevation_Co}
\end{figure}



\begin{figure}
    \centering
    \includegraphics[width=\textwidth]{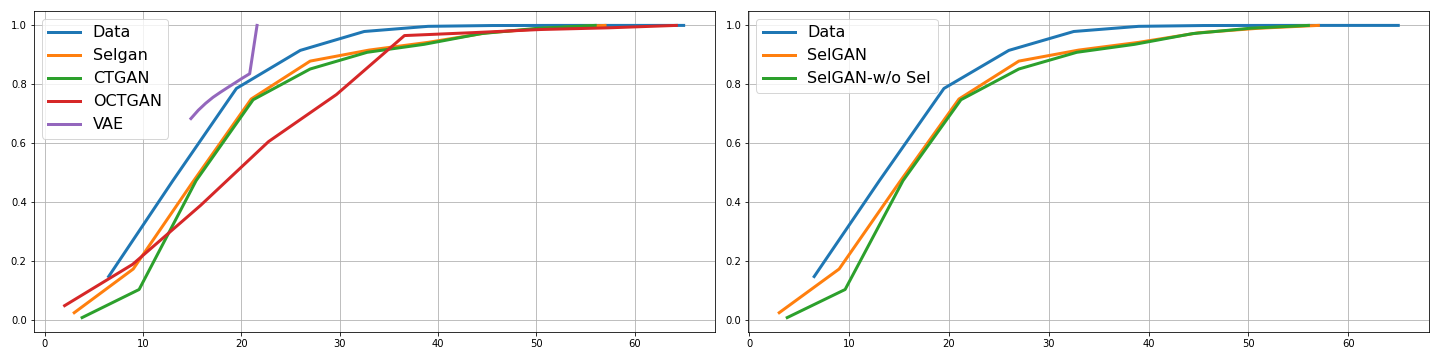}
    \caption{slope in Covertype}
    \label{fig:slope_Co}
\end{figure}


\begin{figure}
    \centering
    \includegraphics[width=\textwidth]{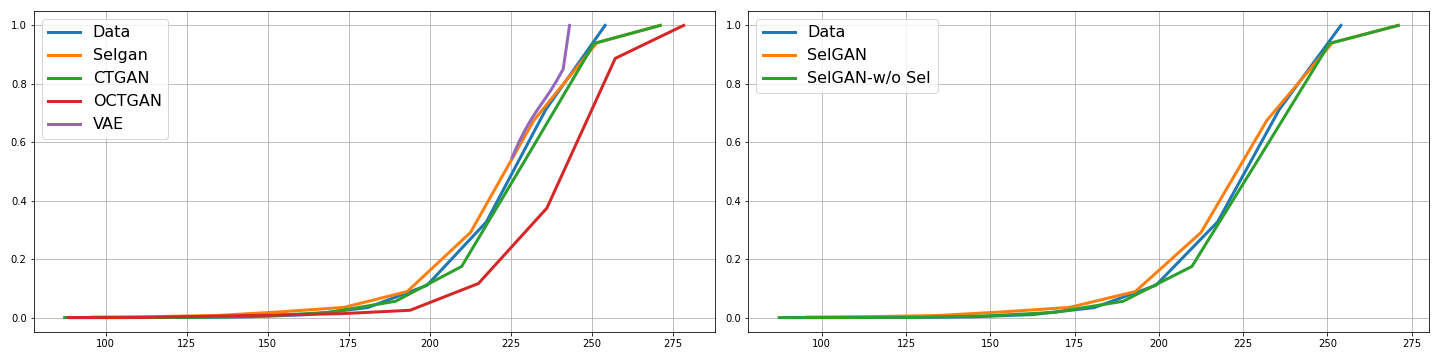}
    \caption{hillshade noon in Covertype}
    \label{fig:hillshade_noon_Co}
\end{figure}


\begin{figure}
    \centering
    \includegraphics[width=\textwidth]{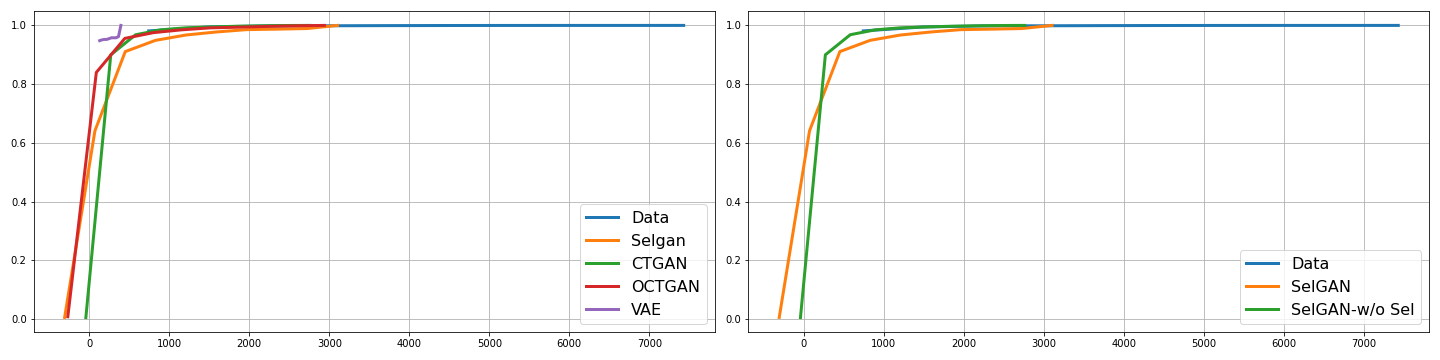}
    \caption{Amount in Credit}
    \label{fig:Amount_Cr}
\end{figure}

\begin{figure}
    \centering
    \includegraphics[width=\textwidth]{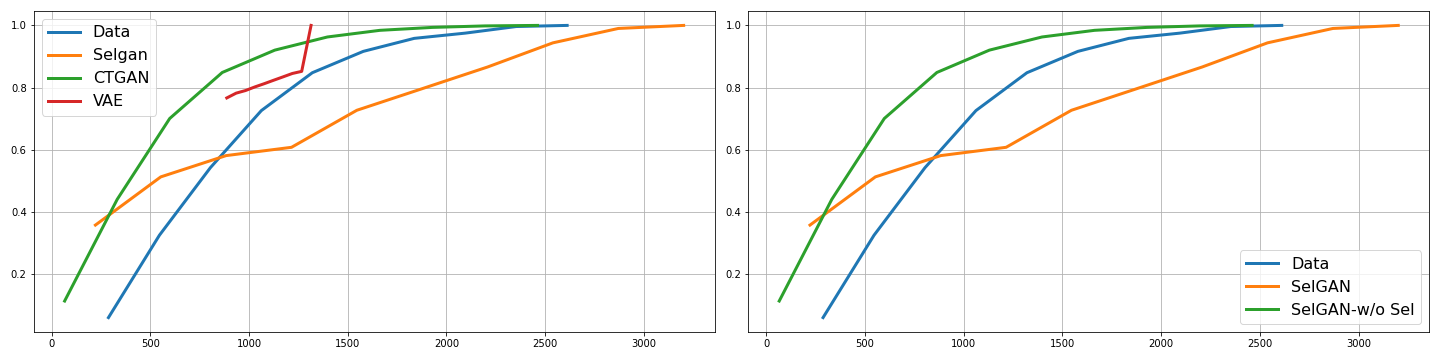}
    \caption{MktDistance in Ticket}
    \label{fig:MktDistance_t}
\end{figure}



\begin{figure}
    \centering
    \includegraphics[width=\textwidth]{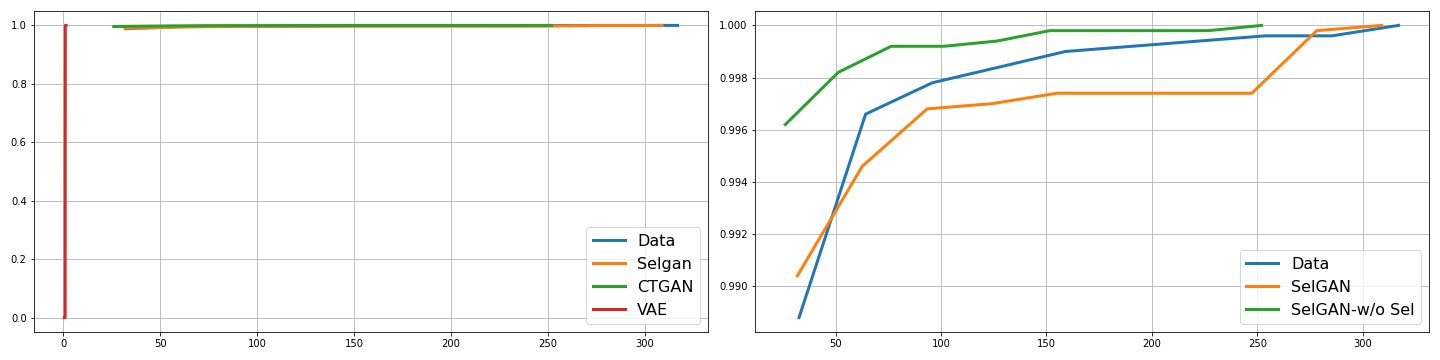}
    \caption{Passengers in Ticket}
    \label{fig:Passengers_t}
\end{figure}


\begin{figure}
    \centering
    \includegraphics[width=\textwidth]{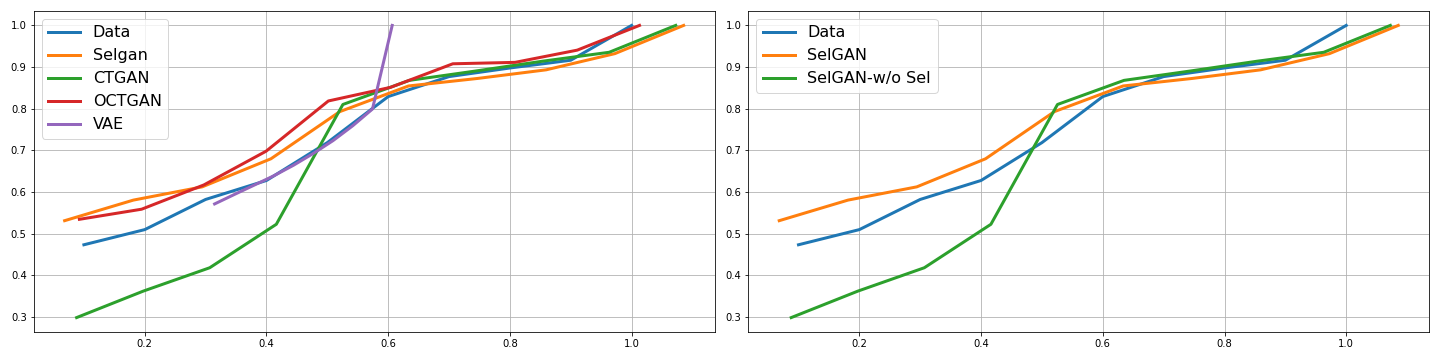}
    \caption{title subjectivity in News}
    \label{fig:title_subjectivity_N}
\end{figure}

\begin{figure}
    \centering
    \includegraphics[width=\textwidth]{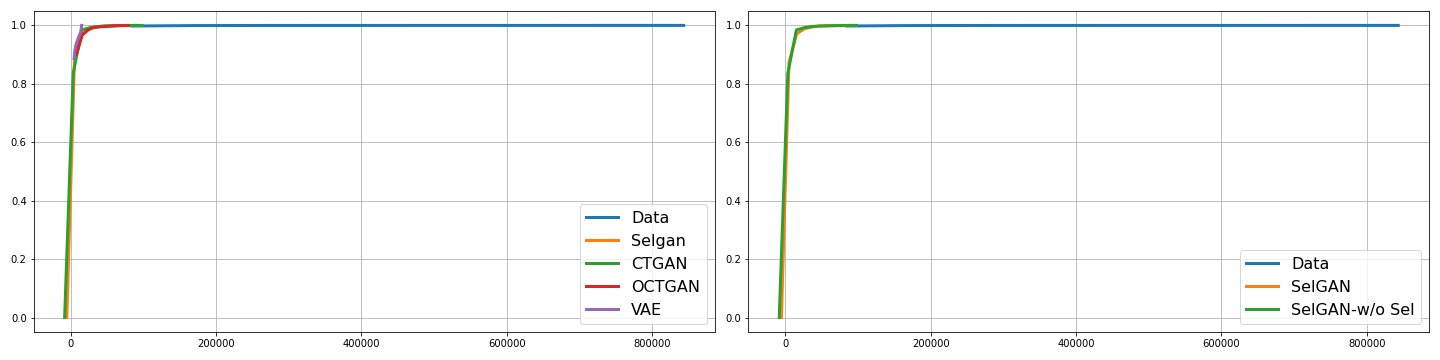}
    \caption{shares in News}
    \label{fig:shares_N}
\end{figure}

\begin{figure}
    \centering
    \includegraphics[width=\textwidth]{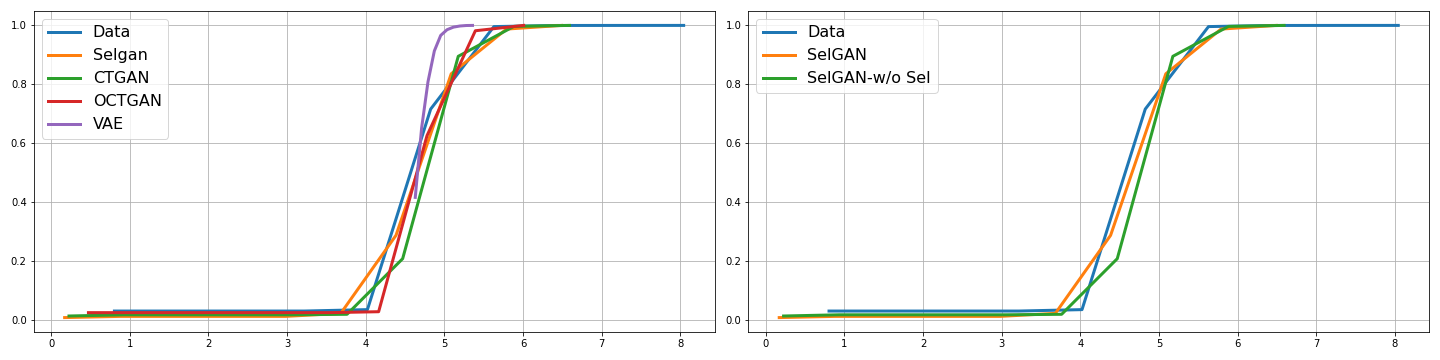}
    \caption{average token length in News}
    \label{fig:average_token_length_N}
\end{figure}

\begin{figure}
    \centering
    \includegraphics[width=\textwidth]{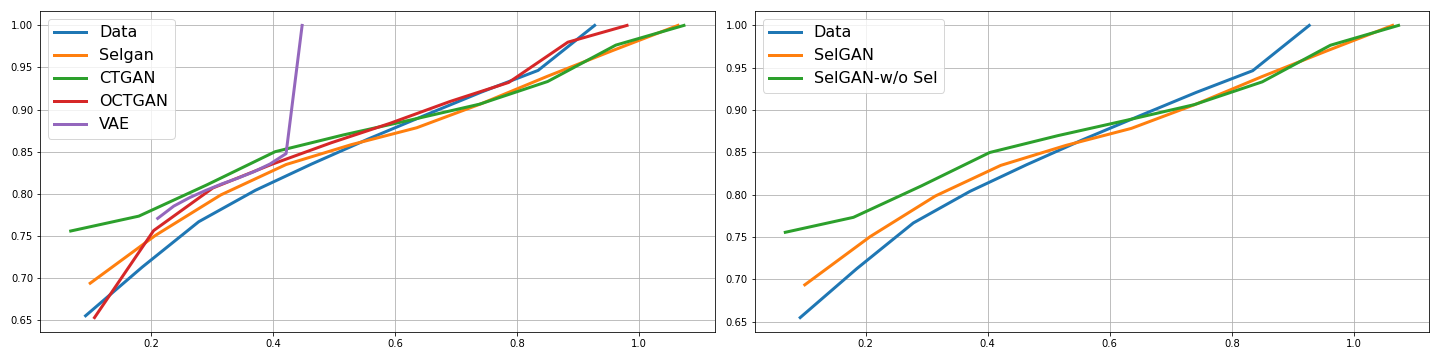}
    \caption{LDA00 in News}
    \label{fig:LDA_00_N}
\end{figure}

\begin{figure}
    \centering
    \includegraphics[width=\textwidth]{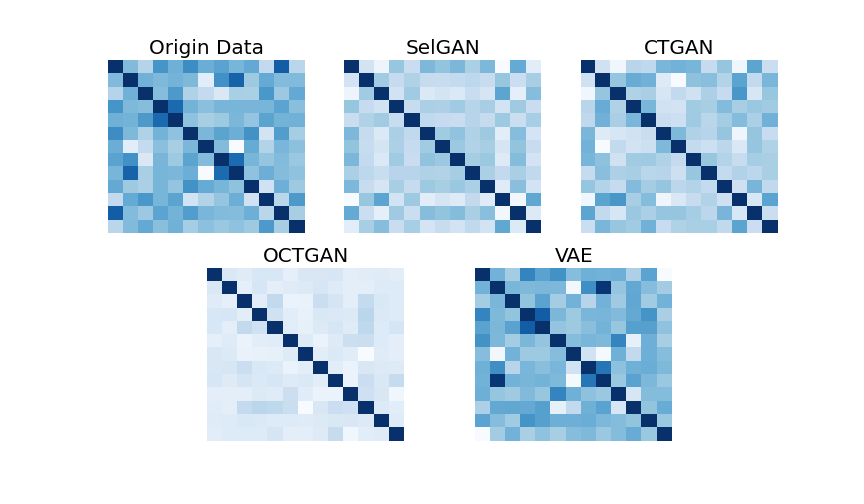}
    \caption{Correlation Heap Map for Covertype}
    \label{fig:Corr_co}
\end{figure}

\begin{figure}
    \centering
    \includegraphics[width=\textwidth]{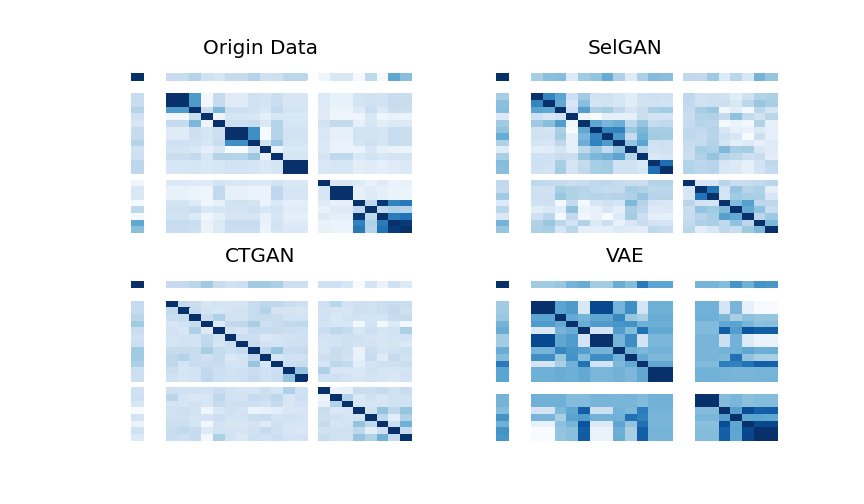}
    \caption{Correlation Heap Map for Ticket}
    \label{fig:Corr_t}
\end{figure}

\begin{figure}[htbp]
    \centering
    \includegraphics[width=\textwidth]{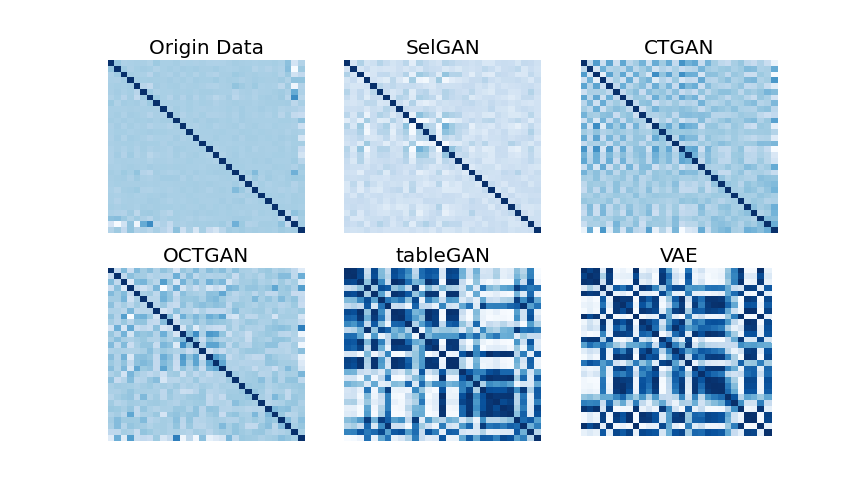}
    \caption{Correlation Heap Map for Credit}
    \label{fig:Corr_cr}
\end{figure} 

\addtocontents{toc}{\vspace{2em}}  
\backmatter

\label{Bibliography}
\lhead{\emph{Bibliography}}  
\bibliographystyle{unsrtnat}  
\bibliography{Bibliography}  

\end{document}